\documentclass{article}

     \usepackage[preprint,nonatbib]{neurips_2020}

\usepackage[utf8]{inputenc} 
\usepackage[T1]{fontenc}    
\usepackage{hyperref,floatrow,blindtext} 
\newfloatcommand{capbtabbox}{table}[][\FBwidth]      
\usepackage{url}            
\usepackage{booktabs,xcolor}       
\usepackage{amsfonts}       
\usepackage{nicefrac}       
\usepackage{microtype}      
\usepackage{multirow,cite}
\usepackage{subcaption,mathtools}
\usepackage{enumerate}
\usepackage{amsmath,amsthm,amssymb}
\usepackage{thm-restate}
\theoremstyle{definition}
\newtheorem{defn}{Definition}[]

\usepackage{pgfplots}
\pgfplotsset{width=10cm,compat=1.9}
\usepgfplotslibrary{external}

\usepackage{authblk}        

\title{Protect, Show, Attend and Tell: Empowering Image Captioning Models with Ownership Protection}

\author[1]{Jian Han Lim}
\author[1]{Chee Seng Chan}
\author[2]{Kam Woh Ng}
\author[2]{Lixin Fan}
\author[3]{Qiang Yang}
\affil[1]{Universiti Malaya, Kuala Lumpur, Malaysia}
\affil[2]{WeBank AI Lab, Shenzhen, China}
\affil[3]{Hong Kong University Science and Technology, Hong Kong}

\begin{document}

\maketitle

\begin{abstract}
By and large, existing Intellectual Property (IP) protection on deep neural networks typically i) focus on image classification task only, and ii) follow a standard digital watermarking framework that was conventionally used to protect the ownership of multimedia and video content. This paper demonstrates that thecurrent digital watermarking framework is insufficient to protect image captioning tasks that are often regarded as one of the frontiers AI problems. As a remedy, this paper studies and proposes two different embedding schemes in the hidden memory state of a recurrent neural network to protect the image captioning model. From empirical points, we prove that a forged key will yield an unusable image captioning model, defeating the purpose of infringement. To the best of our knowledge, this work is the first to propose ownership protection on image captioning task. Also, extensive experiments show that the proposed method does not compromise the original image captioning performance on all common captioning metrics on Flickr30k and MS-COCO datasets, and at the same time it is able to withstand both removal and ambiguity attacks. Code is available at \url{https://github.com/jianhanlim/ipr-imagecaptioning} 
\end{abstract}

\section{Introduction}

Recent advances in deep neural networks (DNN) had significantly improved the overall model performance in multiple artificial intelligence (AI) domains, for example, natural language processing, computer vision, gaming, etc. As a result of this, it has enabled a growing number of AI start-ups and companies to offer their DNN solutions in terms of Software as a Service (SaaS). As such, the protection of the Intellectual Property (IP) of DNN has become a necessity in order to protect the model against IP infringement to preserve the owner's competitive advantage in an open market.

For the past few years, IP protection on DNN \cite{ucida2017,chen2019deepmarks,zhang2018protecting,adi2018turning,guo2018watermarking,le2019adversarial,darvish2019deepsigns,fan2019rethinking} has been a significant research area. Ideally, the goal is the IP protection solution should not degrade the performance of the original model, and at the same time, it must also be resilient against both ambiguity and removal attacks. Although all these existing solutions have achieved this goal, it is unsatisfactory in our view as we found out that all existing DNN watermarking methods have been i) following a standard digital watermarking framework that was conventionally used to protect the ownership of multimedia and video content, and ii) focusing on DNNs for classification tasks that map images to labels and DNNs for other tasks are forgotten such as image captioning that map images to texts.

\begin{figure}[t]
\scriptsize
\centering
  \includegraphics[width=1\textwidth]{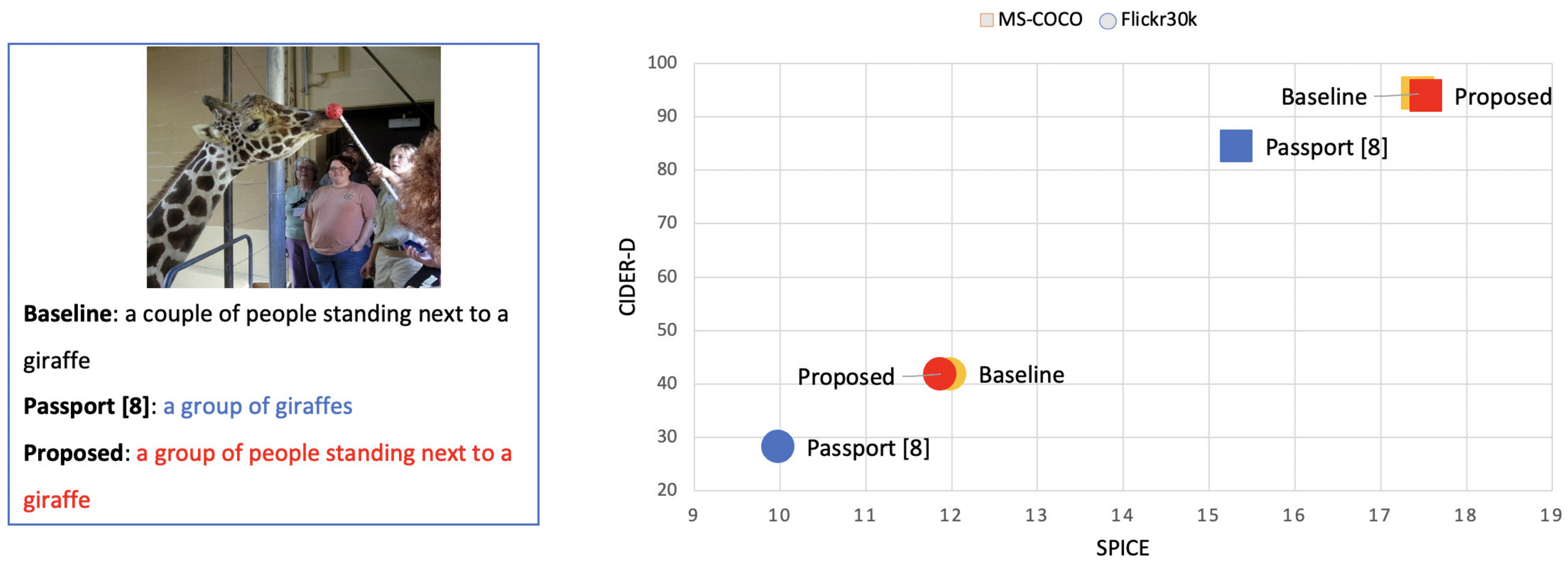}
  \caption{Comparison of the baseline model (in yellow), passport model \cite{fan2019rethinking} (in blue), and proposed model (in red) on two public datasets. Best view in colour.}
  \label{fig:pic1}
\end{figure}

A natural question is then why not directly apply existing watermarking methods \cite{ucida2017,fan2019rethinking} designed for the classification DNNs to watermark the DNNs in image captioning. Unfortunately, it is not the case for the white-box watermarking methods. The obstacles lie in several fundamental differences between these two kinds of DNNs. First, DNNs for classification output a label. In contrast, DNNs for image captioning output a sentence. Second, classification is about finding the decision boundaries among different classes, whereas image captioning is not only to understand the image content in depth beyond category or attribute levels but also to connect its interpretation with a language model to create a natural sentence \cite{xu2015show}. We demonstrated in Figure \ref{fig:pic1}, Figure \ref{fig:qualitative-passport}, and Sect. \ref{cnnwatermark} that a recent digital watermarking framework \cite{fan2019rethinking} that used to protect deep-based classification model is insufficient to protect image captioning models. Figure \ref{fig:pic1} shows that \cite{fan2019rethinking} is insufficient to be deployed to protect the IP of the image captioning task as compared to our proposed model, against the baseline model. For instance, it can be noticed that the caption generated by \cite{fan2019rethinking} is incomplete and incorrect, while the caption generated from our model is very near to the baseline model. In addition, the CIDEr-D and SPICE score of our model is very near to the baseline as compared to \cite{fan2019rethinking} on MS-COCO and Flickr30k datasets. 

As a remedy, this paper proposes a novel embedding framework that consists of two different embedding schemes to embed a unique secret key into the recurrent neural network (RNN) cell \cite{hochreiter1997long} to protect the image captioning model against various attacks. Specifically, we show that embed a secret key into the hidden memory state of an RNN is the best choice for the image captioning task such that a forged key will immediately yield an unusable image captioning model in terms of poor quality outputs, defeating the purpose of infringement.

On the one hand, our solution bears a similarity to digital watermarking - they both embed certain digital entities into models during training sessions. In terms of IP protection, however, embedded watermarks only enable the verification of the ownership of models. One has to rely on government investigation and enforcement actions to discourage IP infringement. Whether this kind of approach can provide {\it reliable}, {\it timely}, and {\it cost-effective} juridical protection remains questionable.  On the other hand, our key-protected models will not function normally unless the valid key is provided, thus immediately preventing the unlawful usages of the models with no extra costs. Indeed, we regard this proactive protection as the most prominent advantage of our solution over digital watermarking. For instance, in Fig. \ref{fig:amb-1}, the protected model with a valid key demonstrated almost identical performance as that of the original model, in contrast the same model presented with a forged key has a huge performance drop in all metric scores.

The contributions are threefold: i) We renovate the paradigm of digital watermarking based neural network IP protection, by proposing a key-based strategy that provides \textit{reliable, preventive} and \textit{timely} IP protection (Sect. \ref{sect:formulation}) at virtually \textit{no extra cost} (Sect. \ref{complexity}) for image captioning task.  ii) This paper formulates the problem and proposes a generic solution as well as concrete implementation schemes that embed a unique key into RNN models through the hidden memory state (Section \ref{approach}; Fig. \ref{fig:keyembed}b). We prove that a forged key will yield a useless image model. Also, we empirically show the effectiveness of our approach against various attacks and prove the ownership of the model (Sect. \ref{experiment}); and iii) To the best of our knowledge, we are the first to propose IP protection on image captioning model and we demonstrated that the proposed method does not compromise the original image captioning performance on all common captioning metrics on Flickr30k and MS-COCO datasets (Table \ref{tab:baseline}). 

\section{Related Work}
Conventionally, digital watermarks were extensively used in protecting the ownership of multimedia contents, including images, videos, audio, or functional designs. It is a process of embedding a marker into the content and subsequently using it to verify the ownership. In deep learning, the IP protection on the models can be categorized into i) white-box based solution \cite{ucida2017,chen2019deepmarks}, ii) black-box based solution \cite{zhang2018protecting,adi2018turning,guo2018watermarking,le2019adversarial,quan2020watermarking} or iii) a combinatorial of both white and black based solution \cite{darvish2019deepsigns,fan2019rethinking}. 

The first work that introduced digital watermarks for DNN was proposed by Uchida et al.\cite{ucida2017}, where the authors embedded a watermark into the weights parameters via parameter regularizer during the training as white-box protection. For verification, owners are required to access the model parameters to extract the watermark. To remedy this issue,  \cite{zhang2018protecting,adi2018turning,guo2018watermarking,le2019adversarial,quan2020watermarking} proposed digital watermarks in a black-box setting. In this setting, a set of trigger set images is generated as random image and label pairs. During training, the feature distributions of those images are distant from the labeled training samples. During verification, the trigger set watermark can be extracted remotely without the need to access the model weights. For example, Zhang et al. \cite{zhang2018protecting} introduced three different key generations which are content-based, noise-based, and unrelated-based images respectively. Adi et al. \cite{adi2018turning} proposed a watermarking method similar to \cite{zhang2018protecting} but their main contribution is the model verification. While Merrer et al. \cite{le2019adversarial} proposed to use adversarial examples as the watermark key set to modify the model decision boundary. Quan et al. \cite{quan2020watermarking} aimed to develop a black-box watermarking method in images to images tasks such as image denoising and super-resolution by exploiting the overparameterization of the model.

Recently, \cite{darvish2019deepsigns, fan2019rethinking} presented a watermarking framework that works in both white-box and black-box settings. Rouhani et al. \cite{darvish2019deepsigns} embedded watermark in activation of selected layers of the DNN by integrating two additional regularization loss terms, binary cross-entropy loss, and Gaussian Mixed Model (GMM) agent loss. It is robust against pruning, fine-tuning, and overwriting attack but require more computation. The work that most closer to us is Fan et al. \cite{fan2019rethinking} added special ``passport" layers into the DNN model to enable ownership verification. With a forged passport, the performance of the model will significantly deteriorate. This design relies on the secrecy of passport layer weights that requires the owner to keep the passport layer weights secret from the attacker. However, empirically, we demonstrated that \cite{fan2019rethinking} does not able to protect the image captioning model effectively.

\section{Approach}
\label{approach}
Our image captioning framework of interest is a simplified variant of the \emph{Show, Attend and Tell} model \cite{xu2015show}. It is a popular framework that forms the basis for subsequent state-of-the-art works on image captioning \cite{anderson2018bottom,bernardi2016automatic,rennie2017self,karpathy2015deep,ding2019image,he2019image,xiao2019dense,wang2020learning,ji2021divergent}. It follows the encoder-decoder framework, where a convolutional neural network (CNN) is used to encode an image into a fixed-size representation, and the long short-term memory (LSTM) is employed to generate the captions. 

Given an image $I$, it is encoded into a fixed-size feature vector using CNN as followed:
\begin{equation}
    X = f_c(I)
\end{equation}
where $f_c(\cdot)$ represents CNN encoder in our models, $X$ is the image feature vector.

In the decoder, the image feature vector $X$ is fed into LSTM at each time step $t$ to output the probability of next word $S_t$ as:
\begin{equation}
h_{t} = LSTM(X, h_{t-1}, m_{t-1})
\end{equation}
\begin{equation}
p(S_{t}|S_{0},\cdots,S_{t-1},I) = F_1(h_{t})
\end{equation}
where $h_{t-1}$ is the previous LSTM's hidden state, $m_{t-1}$ is the previous memory cell, $F_1(\cdot)$ is a nonlinear function that outputs the probability of $S_{t}$, $p$ is the probability of next word $S_{t}$ with image $I$ and previous words $S_{0},\cdots,S_{t-1}$.

Unless otherwise stated, our models are trained under the maximum likelihood estimation (MLE) framework, where the probability of generating a correct caption of length $T$ with tokens $\left\{S_0,\: \cdots\:, \:S_{T-1}\right\}$ for an image $I$ is directly maximized:
\begin{equation}\label{eq: log-likelihood}
    \log{p}\left( S\,|\,I \right)=\sum_{t\,=\,0}^T\log{p}\left( S_t\,|\,I,\; S_{0\,:\,t-1}, \;c_t \right)
\end{equation}
\noindent where $t$ is the time step, $p\left( S_t\,|\,I,\; S_{0\::\:t-1},\; c_t \right)$ is the probability of generating a word given an image $I$, previous words $S_{0\::\:t-1}$, and context vector $c_t$.

\begin{figure*}
\centering
   \begin{subfigure}[t]{0.45\textwidth}
    \centering
    \includegraphics[width=0.8\textwidth]{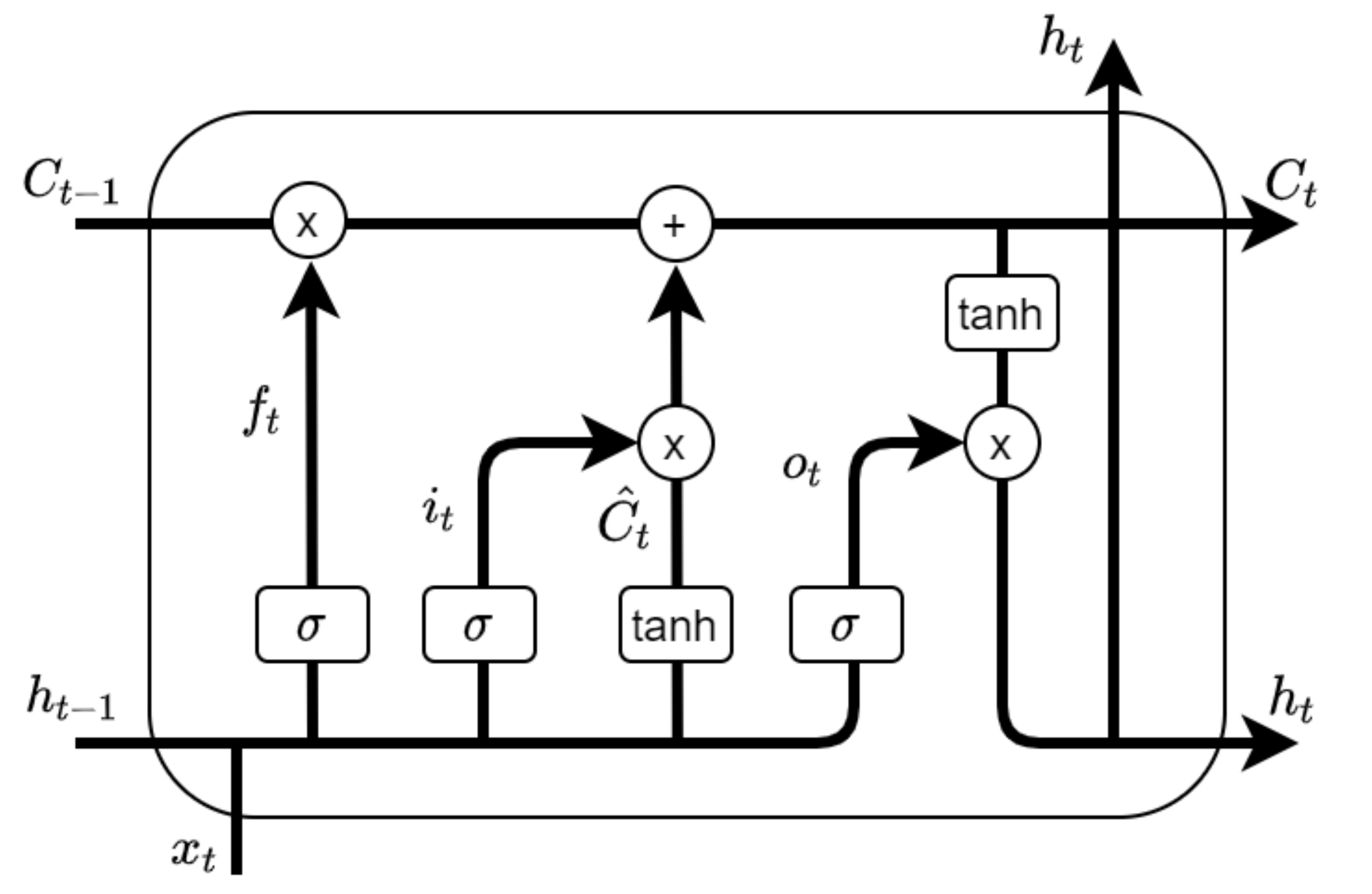}
    \\(a) Original LSTM Cell
  \end{subfigure}
  \begin{subfigure}[t]{0.45\textwidth}
    \centering
    \includegraphics[width=0.85\textwidth]{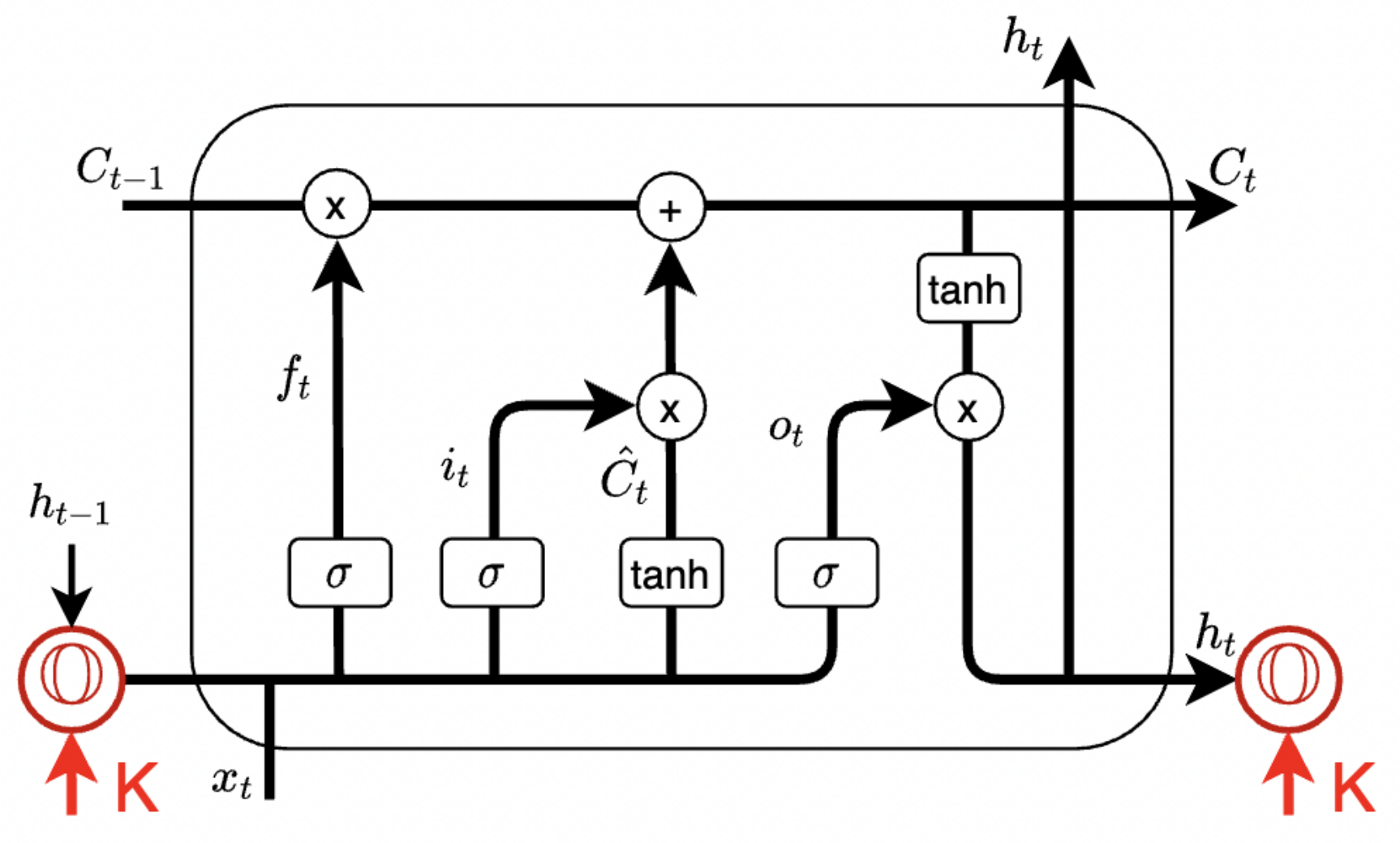}
    \\(b) LSTM Cell with Secret Key Embedding
  \end{subfigure}
  \caption{An overview of our approach. (a) The original LSTM Cell and (b) LSTM Cell with key embedding operation \(\mathbb{O}\) $\in$ \{$M_\oplus$, $M_\otimes$\} (see Section \ref{eoperation})}
  \label{fig:keyembed}
\end{figure*}

\subsection{Problem formulation}\label{sect:formulation}
Let $\mathcal{N}$ denote an image captioning model to be protected by a \textit{secret} key $k$, after a training process, the image captioning model embedded with the key is denoted by $\mathcal{N}[K]$ as shown in Figure \ref{fig:keyembed}. The inference of such a protected model can be characterized as a process $M$ that \textit{modifies} the model behavior according to the running-time key $l$: 
\begin{align}\label{eq:dpp-formula}
M( \mathcal{N}[K], l) = \left\{ \begin{array}{cc}
\mathcal{M}_K, & \text{if }l=K,  \\
\mathcal{M}_{\bar{K}}, & \text{otherwise},  \\
\end{array} \right.
\end{align}
in which $\mathcal{M}_K$ is the network performance with key correctly verified, and $\mathcal{M}_{\bar{K}}$ is the performance with the incorrect key i.e., $\bar{K} \neq K$. 

The properties  of  $M( \mathcal{N}[K], l)$ defined below are desired for the sake of IP protection: 
\begin{defn}
	If $l=K$, the performance $ \mathcal{M}_K $ should be as close as possible to that of the original network $\mathcal{N}$. Specifically, if the performance \textit{inconsistency} between $ \mathcal{M}_K $ and that of $\mathcal{N}$ is smaller than a desired threshold, then the protected network is called \textit{functionality-preserving}.
\end{defn}

\begin{defn}
If $l\neq K$, on the other hand, the performance $\mathcal{M}_{\bar{K}}$  should be as far as possible to that of $\mathcal{M}_{{k}}$. The discrepancy between $\mathcal{M}_{{K}}$ and  $\mathcal{M}_{\bar{K}}$ therefore can be defined as the \textit{protection-strength}. 
\end{defn}

\subsection{Embedding operation}
\label{eoperation}

\begin{figure*}
\centering
   \begin{subfigure}[t]{1.0\textwidth}
    \centering
    \scriptsize
    \includegraphics[width=0.7\textwidth]{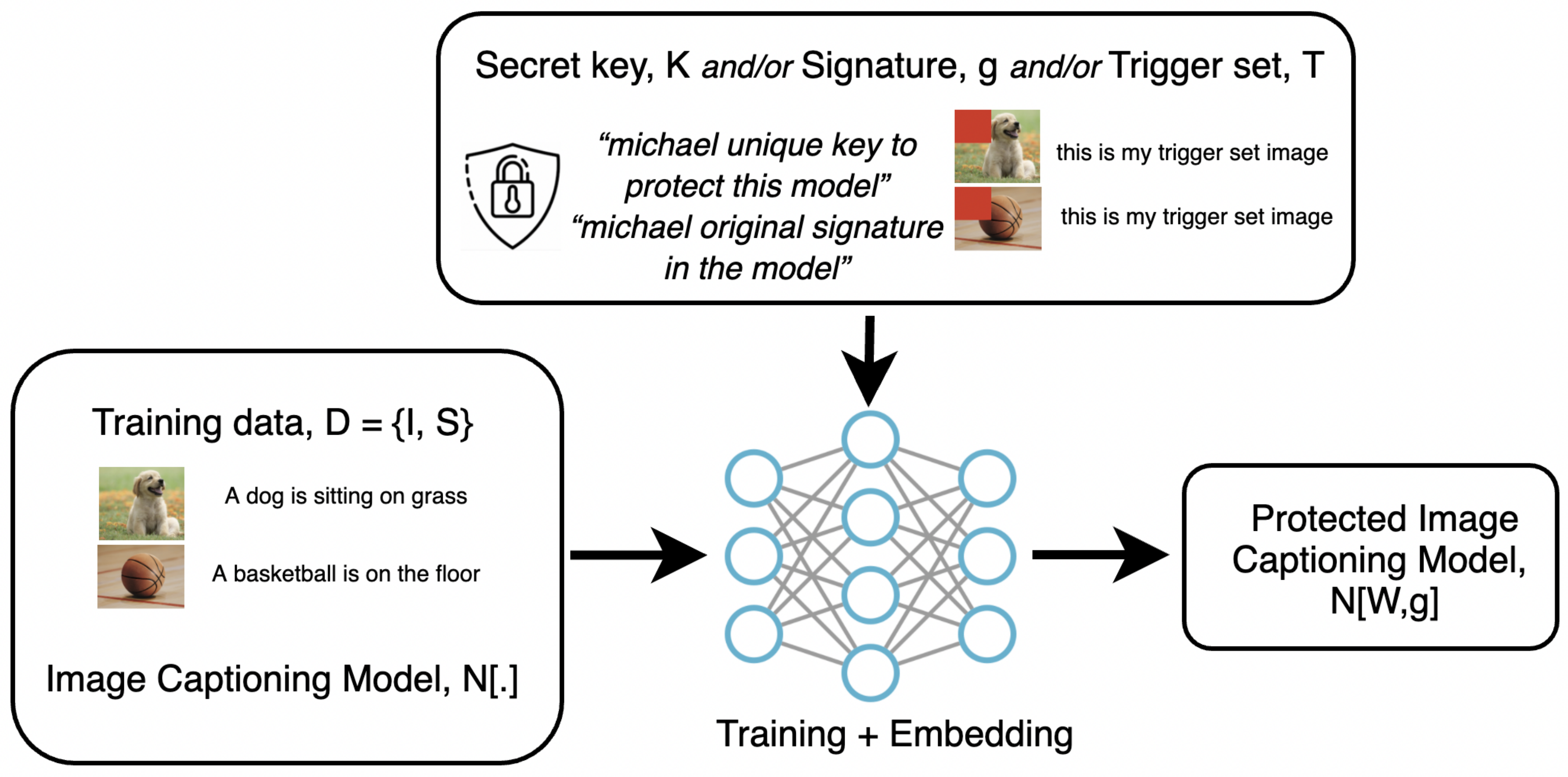}
    \\(a) An embedding process $E_{\mathbb{O}}$, takes inputs training data $\mathbf{D}=\{I, S\}$, secret key $K$ and/or signature $g$ and/or trigger set $T$, model $\mathcal{N}[.]$ to produce protected model $\mathcal{N}[W,g]$.
    \vspace{10pt}
  \end{subfigure}
  \begin{subfigure}[t]{1.0\textwidth}
    \centering
    \scriptsize
    \includegraphics[width=0.7\textwidth]{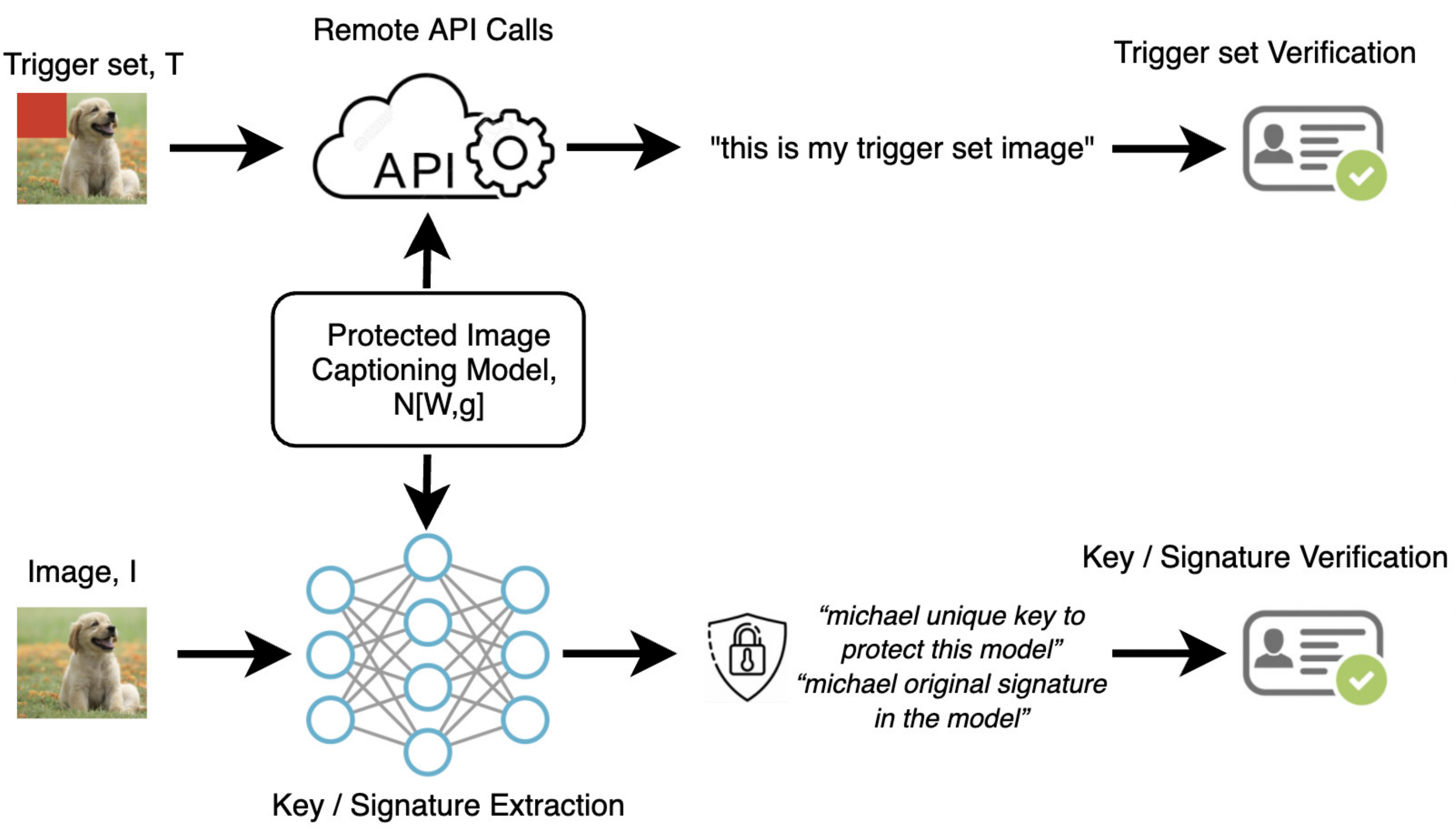}
    \\(b) A verification process $V$ takes as inputs, either an image $I$ or a trigger set $T$, and outputs the result to verify the ownership.
  \end{subfigure}
  \caption{Visual explanation for embedding and verification processes.}
  \label{fig:visualprocess}
\end{figure*}

Figure \ref{fig:visualprocess}a shows the overview of the embedding process. Our \textit{embedding} process can be represented as $E_{\mathbb{O}}\big( \mathbf{D}, \mathbf{g}, \mathcal{N}[.], L  \big) = \mathcal{N}[\mathbf{W,g}]$, is a RNN learning process. It takes inputs \textit{training data} $\mathbf{D}=\{I, S\}$, and optionally signature $\mathbf{g}$, and optimizes the model $\mathcal{N}[\mathbf{W,g}]$ by minimizing the given loss $L$. In this paper, we introduce two different key embedding operations $\mathbb{O}$ which are i) element-wise addition model ($M_\oplus$) or ii) element-wise multiplication model ($M_\otimes$):
\begin{equation}
\mathbb{O}(K, h_{t-1}, e) = 
\begin{cases}
  K \oplus h_{t-1}, & \text{if }
       \!\begin{aligned}[t]
       e = \oplus,
       \end{aligned}
\\
  K \otimes h_{t-1}, & else.
\end{cases}
\end{equation}
where \(k_{f} = \{k_{f,i}\}^{N}_{i=1}\) with $N$ is the size of the hidden state, \(k_{f,i} \in \mathbb{R} :-1 \leq k_{f,i} \leq 1\) and \(k_{b} = \{k_{b,i}\}^{N}_{i=1}\) with \(k_{b,i} \in  \{-1, 1\}\). 

Then, the embedded key $K$ is represented in terms \(k_{b}\) is generated by converting the string provided by owner to a binary vector \(BE\). However, we found that the binary vector for very near alphanumeric, e.g., string A and C, has only a 1-bit difference. Therefore, we proposed a new transformation function $\mathbb{T}$:
\begin{equation}
\mathbb{T}(BE, BC) = BE \otimes BC = k_{b}
\end{equation}
where \(BC\) is a binary vector sampled from value of -1 or 1 according to the seed provided by user, to alleviate this issue.

\subsection{Sign of hidden state as signature}
\label{sign}

In order to further strengthen our model, we follow \cite{fan2019rethinking} to add the sign loss regularization term into the loss function as:
\begin{equation}
L_g(h, G, \gamma) = \sum_{i=1}^{N} max(\gamma - h_{i}g_{i},0)
\end{equation}
where \(G = \{g_{i}\}^{N}_{i=1}\) with \(g_{i} \in \{-1, 1\}\)  consists of the designated binary bits for hidden state \(h\). To enforce the hidden state to have a magnitude greater than \(0\), a hyperparameter \(\gamma\) is introduced into the sign loss. However, one of the main differences of our approach compared to \cite{fan2019rethinking} is our signature is not embedded in the model weights, but it is embedded in the hidden state which is the output of the LSTM cell. This is because we found out that embedding signature in the model weights can be easily attacked with a channel permutation, i.e., change the signature but remains the output of the model. 

\subsection{Ownership verification}
Figure \ref{fig:visualprocess}b shows the overview of the verification process. Suppose an owner tries to verify the ownership of a target model, three verification methods are proposed: 1) $V_1$: Key-based verification; 2) $V_2$: Signature-based verification; and 3) $V_3$: Trigger set verification.

\textbf{$V_1$: Secret key-based verification} - In this verification scheme, there are two different approaches, depending on the secret key is either public or private. Formerly, the trained model and the public key will be provided to the clients. For model inferences, the public key will be required as part of the input to the model to ensure the model performance is preserved. The ownership of the model can be verified directly by the provided key. Latter, a private key is directly embedded into the model. For inference, only image is required as the model input. However, for ownership verification, one has to have access to the model and extract the key from the LSTM cell. 

\textbf{$V_2$: Signature verification} - In this verification scheme, a unique signature is embedded in the sign of the hidden state during the training process via sign loss regularization. To verify the signature, the owner is required to access the trained model. Then, an image will be sent to the model to generate the caption. During inference time, the sign of the hidden state of the LSTM cell will be extracted and compared with our signature to verify the ownership. This binary bits signature can be transformed back to a human-readable string for example the name of the owner.  

\textbf{$V_3$: Trigger set verification} - $V_1$ and $V_2$ are considered as white-box verification, where the owner is required to have access to the model physically in order to verify the ownership. Hence, we introduce trigger set verification that can be conducted remotely via API calls. First, a set of trigger set image-caption pairs\footnote{It is actually a list of data wrongly labeled by purpose} are generated, and then they are used together with the original training samples to train an image captioning model. In this paper, the trigger set images are generated by adding noise (e.g., red color patch) to an original image so that the model is trained to generate the trigger set caption (e.g., this is my trigger set image). For verification, the owner will send the trigger set images to the model and test whether the model returns the trigger set caption.

\section{Experiments}
\label{experiment}

This section presents the experiment results of our approaches in terms of resilience to ambiguity and robustness to removal attacks. Qualitative analysis is also carried out to compare the quality of the caption generated by different approaches. We compare the following models: (i) \textbf{Baseline} is implemented based on the soft attention model as to \cite{xu2015show}, it is an unprotected model. (ii) \textbf{Passport} \cite{fan2019rethinking} is the work that most closer to us that added ``passport" layers into the DNN model to enable ownership verification. (iii) {\boldmath$M_{\oplus}$} is our element-wise addition model presented in Section \ref{eoperation}. (iv) {\boldmath$M_{\otimes}$} is our element-wise multiplication model presented in Section \ref{eoperation}.

We used ResNet-50 \cite{he2016deep} pre-trained on the ImageNet dataset as the encoder. The image features are extracted using ResNet-50 without fully connected layers, resulting in $7 \times 7 \times 2048$ dimensional outputs. For the decoder part, we use LSTM with a dropout rate of 30\%. Both the word embedding and hidden state are set to 512. We set the attention loss factor to 0.01. The LSTM decoder is trained using a learning rate of 1e-4 for 8 epochs and finetune the CNN with a learning rate of 1e-5 up to 20 epochs. The model is trained by cross-entropy loss with a mini-batch size of 32 using Adam \cite{kinga2015method} optimizer with \(\beta_{1}\) set to 0.9, \(\beta_{2}\) set to 0.999, and \(\epsilon\) set to 1e-6. We apply gradient clipping to prevent any gradient to have norm greater than 5.0 to prevent exploding gradients. We repeated all experiments 3 times to get the average performance. The beam size is set as 3 in the inference stage.

\subsection{Dataset and metrics}
We train and evaluate our approaches on the MS-COCO \cite{lin2014microsoft} and Flickr30k \cite{young2014image} datasets, which are widely used for the image captioning task. We followed the widely used split in \cite{karpathy2015deep} for both datasets. MS-COCO contains 113,287 training images with 5 human-annotated captions for each image. The validation and test sets contain 5,000 images each. Flickr30k contains 1,000 images for validation, 1,000 for test, and the rest for training. We truncated captions longer than 20 words and converted all the words into lower case. Fixed vocabulary size of 10,000 is used for both datasets.

We evaluate our approaches using all common metrics in the image captioning task: CIDEr-D \cite{vedantam2015cider}, SPICE \cite{anderson2016spice}, BLEU \cite{papineni2002bleu}, METEOR \cite{banerjee2005meteor}, and ROUGE-L \cite{lin-2004-rouge}. However, CIDEr-D and SPICE have been shown to have a higher correlation with human judgments compared to BLEU and ROUGE \cite{vedantam2015cider,anderson2016spice}, but it is common practice to report all the aforementioned metrics in the image captioning task. 

\begin{table}[t]
\centering
\resizebox{0.95\textwidth}{!}{ 
\renewcommand*{\arraystretch}{1.5}
\Huge
\begin{tabular}{|c|cccccccc|cccccccc|}
\hline
\multirow{2}{*}{Methods} & \multicolumn{8}{c|}{MS-COCO} & \multicolumn{8}{c|}{Flickr30k}\\ \cline{2-17} 
  & B-1 & B-2 & B-3 & B-4 & M & R & C & S & B-1 & B-2 & B-3 & B-4 & M & R & C & S \\ \hline
Baseline & 72.14 & 55.70 & 41.86 & 31.14 & 24.18 & 52.92 & 94.30 & 17.44 & 63.40 & 45.18 & 31.68 & 21.90 & 18.04 & 44.30 & 41.80 & 11.98 \\
\hline
Passport \cite{fan2019rethinking} & 68.50 & 53.30 & 38.41 & 29.12 & 21.03 & 48.80 & 84.45 & 15.32 & 48.30 & 38.23 & 26.21 & 17.88 & 15.02 & 32.25 & 28.22 & 9.98 \\  
$M_{\oplus}$ &\bf{72.53}&\bf{56.07}&\bf{42.03}&\bf{30.97}&\bf{24.00}&\bf{52.90}&$^{*}$91.40&$^{*}$17.13&\bf{62.43}&\bf{44.40}&\bf{30.90}&\bf{21.13}&$^{*}$17.53&\bf{43.63}&$^{*}$40.07&$^{*}$11.57\\
$M_{\otimes}$ &$^{*}$72.47&$^{*}$56.03&$^{*}$41.97&$^{*}$30.90&$^{*}$23.97&\bf{52.90}&\bf{91.60}&\bf{17.17}&$^{*}$62.30&$^{*}$44.07&$^{*}$30.73&$^{*}$21.10&\bf{17.63}&$^{*}$43.53&\bf{40.17}&\bf{11.67}\\
\hline
\end{tabular}
}
  \caption{Comparison between our approaches ($M_{\oplus}$,$M_{\otimes}$) with baseline and Passport \cite{fan2019rethinking} on MS-COCO and Flickr30k datasets, across 5 common metrics where B-N, M, R, C, and S are BLEU-N, METEOR, ROUGE-L, CIDEr-D, and SPICE scores. {\bf BOLD} is the best result and $^{*}$ is the second best result.}
  \label{tab:baseline}
\end{table}

\begin{figure}[h]
\scriptsize
\centering
  \begin{subfigure}[t]{0.3\textwidth}
    \centering
    \includegraphics[width=3cm,height=2cm]{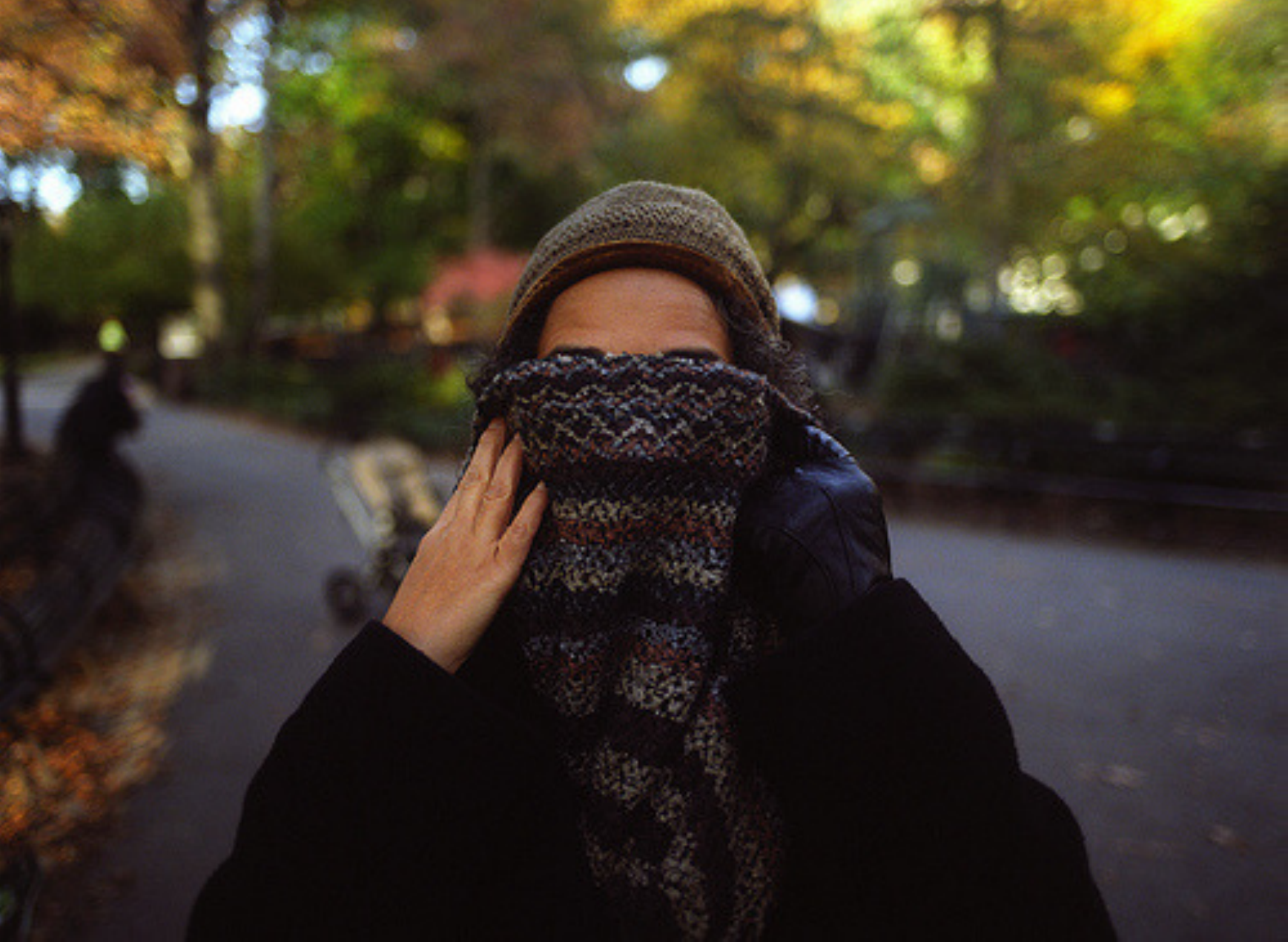}
    
    \raggedright
    \textbf{(a)} {\bf a woman wearing a hat and a scarf.} \\
    (b) {\color{blue} a woman in a scarf is standing.}\\
    (c) a woman in a hat is standing.\\
    (d) {\color{red} a woman.}\\
    
  \end{subfigure}
  \begin{subfigure}[t]{0.35\textwidth}
    \centering
    \includegraphics[width=3cm,height=2cm]{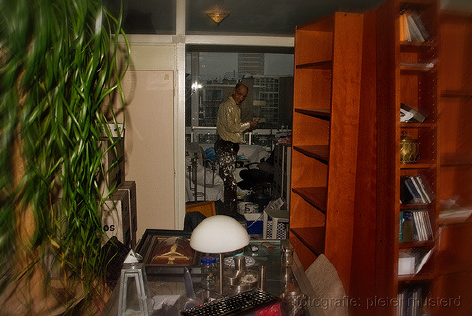}
    
    \raggedright
    \textbf{(a)} {\bf a man in a white shirt is standing in a room.} \\
    (b) {\color{blue} a man in a white shirt is standing in a room.}\\
    (c) a man in a white shirt is standing in a room.\\
    (d) {\color{red} a man in room.}\\
    
  \end{subfigure}
  \begin{subfigure}[t]{0.33\textwidth}
    \centering
    \includegraphics[width=3cm,height=2cm]{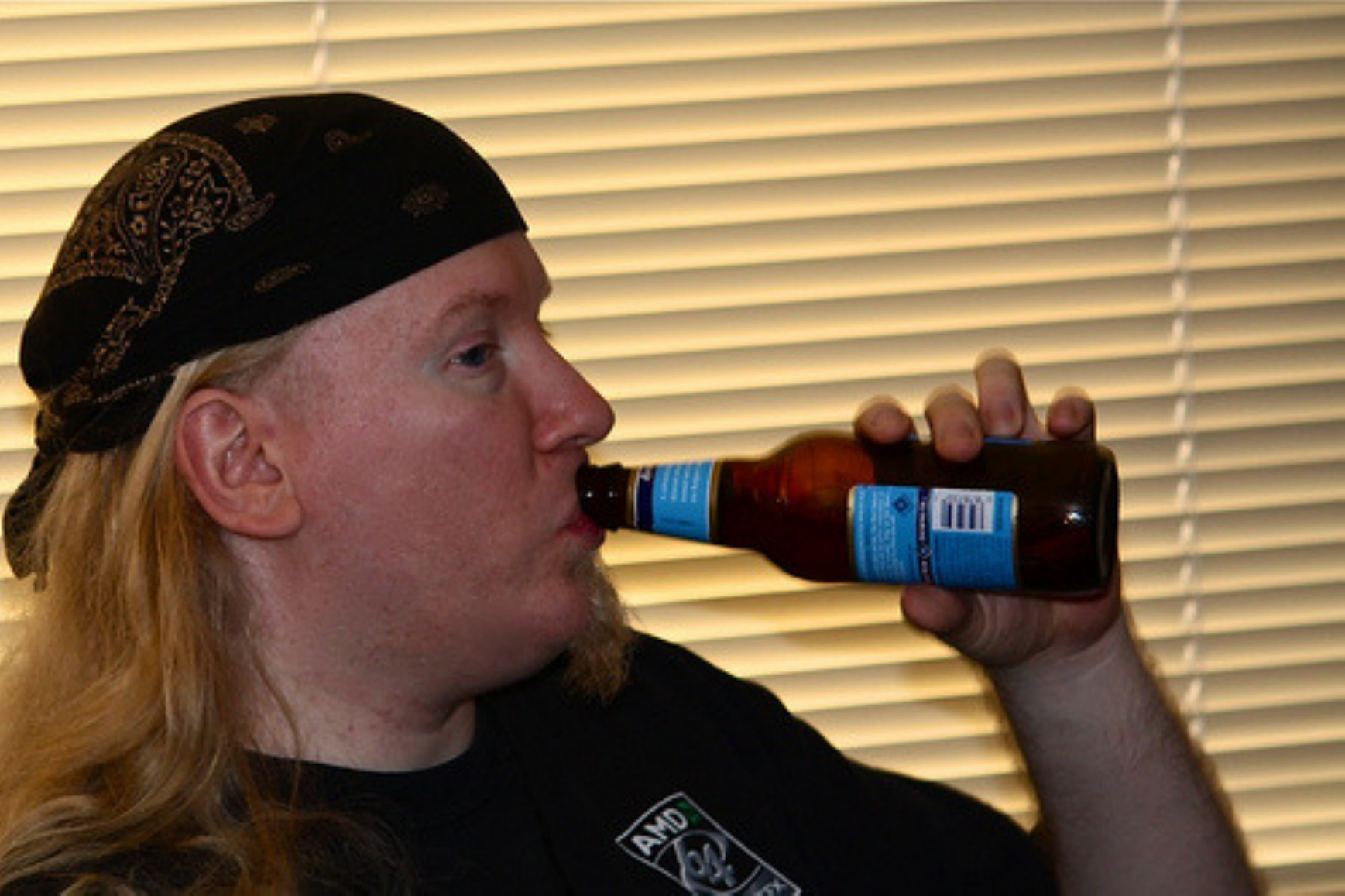}
    
    \raggedright
    \textbf{(a)} {\bf a man drinking a drink.} \\
    (b) {\color{blue} a man is drinking beer.}\\
    (c) a man is drinking a beer.\\
    (d) {\color{red} a man is drinking.}\\
    
  \end{subfigure}
  \vspace{10pt}
  
  \begin{subfigure}[t]{0.3\textwidth}
    \centering
    \includegraphics[width=3cm,height=2cm]{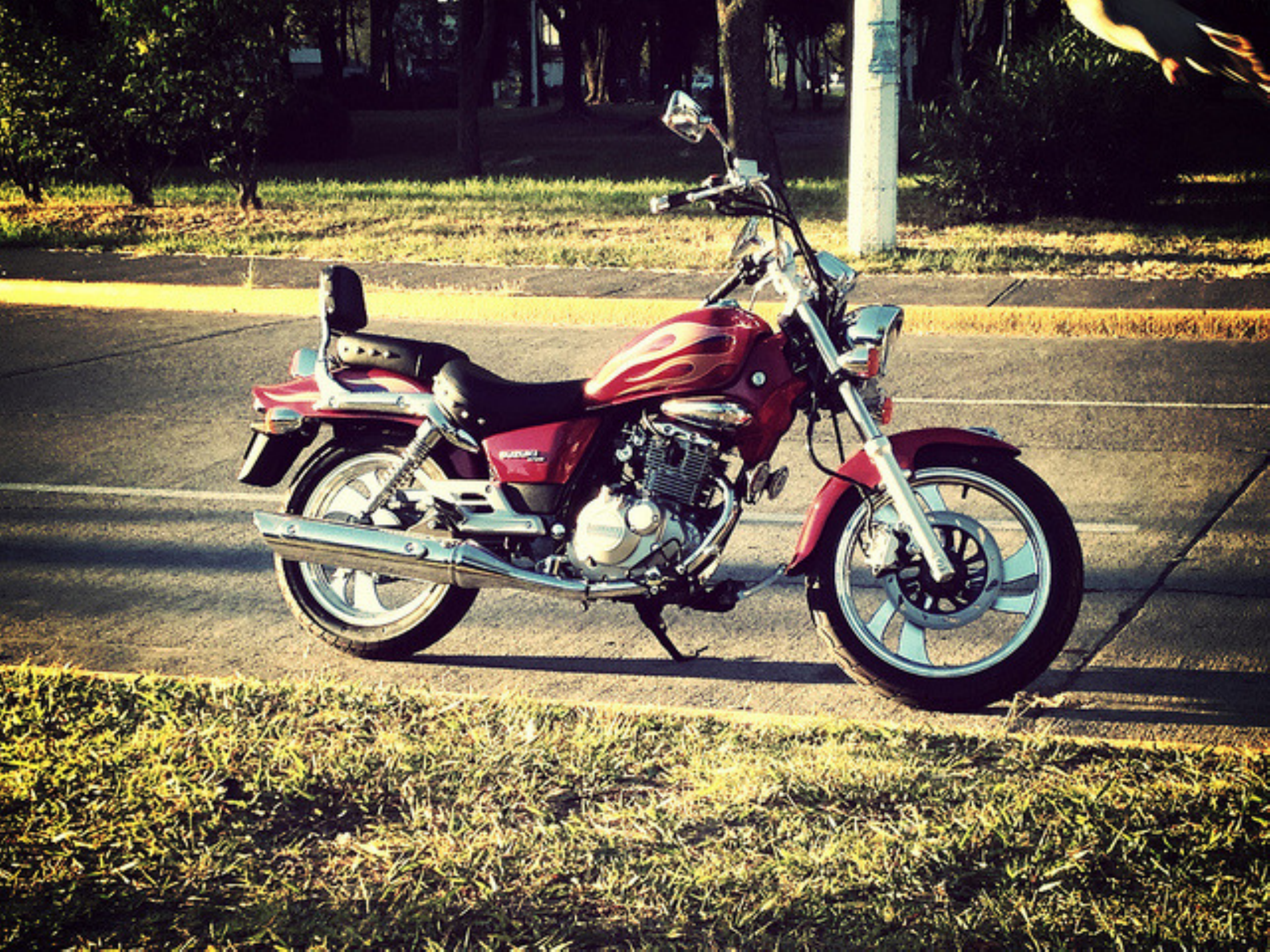}
    
    \raggedright
    \textbf{(a)} {\bf a motorcycle parked on the side of a road.} \\
    (b) {\color{blue} a motorcycle parked on the side of a road.}\\
    (c) a motorcycle parked on the side of a road.\\
    (d) {\color{red} a motorcycle on the road.}\\
    
  \end{subfigure}
  \begin{subfigure}[t]{0.33\textwidth}
    \centering
    \includegraphics[width=3cm,height=2cm]{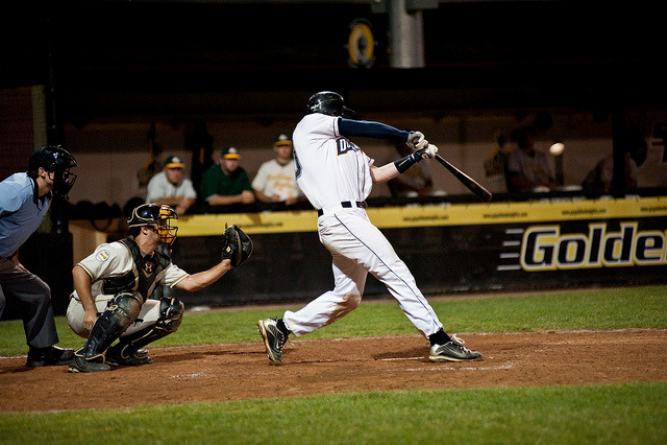}
    
    \raggedright
    \textbf{(a)} {\bf a baseball player swinging a bat at a ball.} \\
    (b) {\color{blue} a baseball player swinging a bat at a baseball game.}\\
    (c) a baseball player swinging a bat on a field.\\
    (d) {\color{red} a baseball player swinging a bat.}\\
    
  \end{subfigure}
  \begin{subfigure}[t]{0.33\textwidth}
    \centering
    \includegraphics[width=3cm,height=2cm]{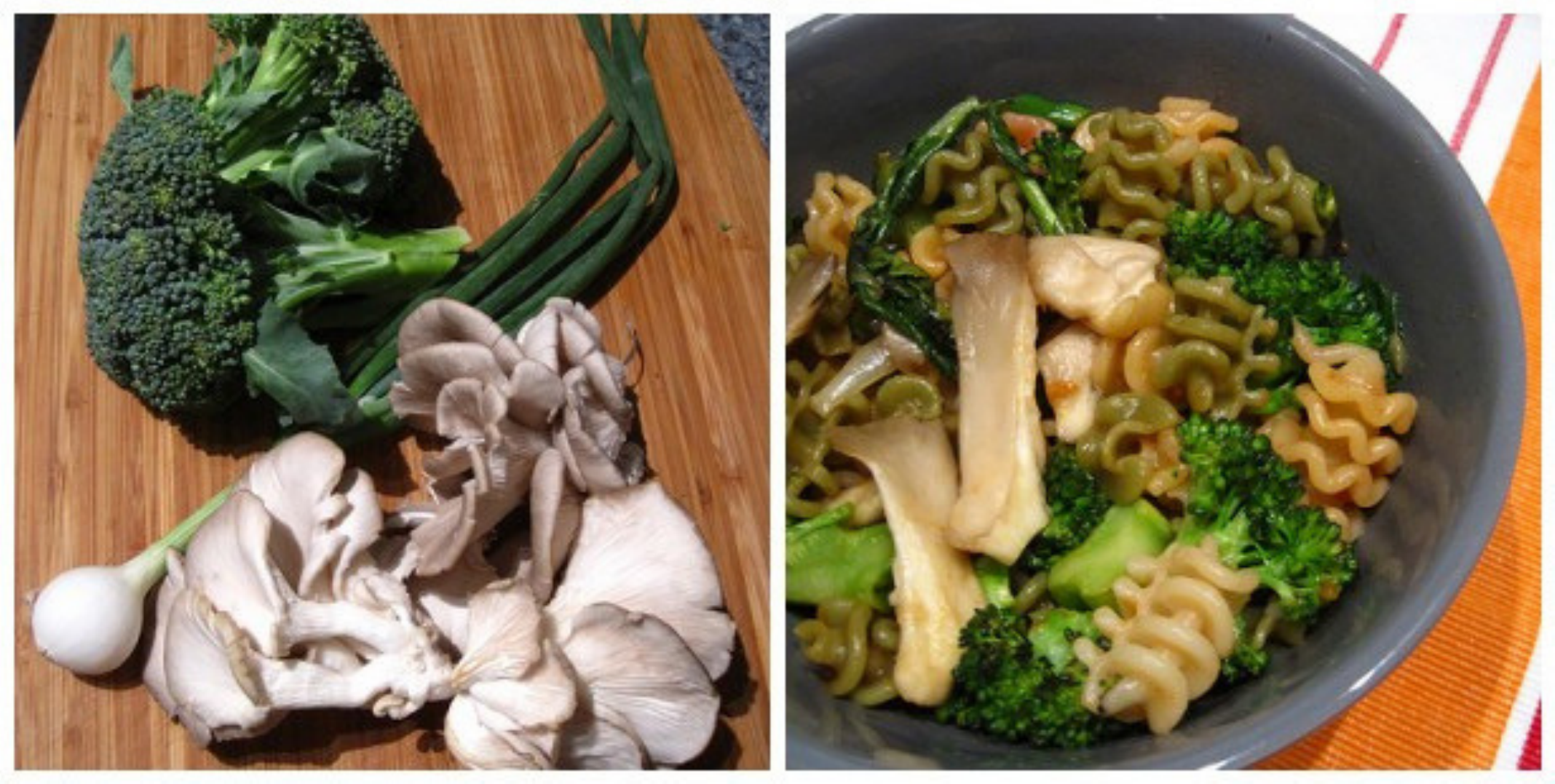}
    
    \raggedright
    \textbf{(a)} {\bf a close up of a plate of food with broccoli.} \\
    (b) {\color{blue} a close up of a bowl of broccoli.}\\
    (c) a bowl of food with broccoli.\\
    (d) {\color{red} a couple of broccoli and broccoli on a table.}\\
    
  \end{subfigure}
  
  \caption{Comparison of captions generated by (a) Baseline, (b) $M_{\oplus}$ (c) $M_{\otimes}$, and (d) Passport \cite{fan2019rethinking}. The first row is the images from Flickr30k. The second row is the images from the MS-COCO dataset. It is noticed that the quality of the captions generated by our models is very close to the baseline.}
  \label{fig:qualitative-passport}
\end{figure}

\begin{figure}[h]
\scriptsize
\centering
  \begin{subfigure}[t]{0.33\textwidth}
    \centering
    \includegraphics[width=3cm,height=2cm]{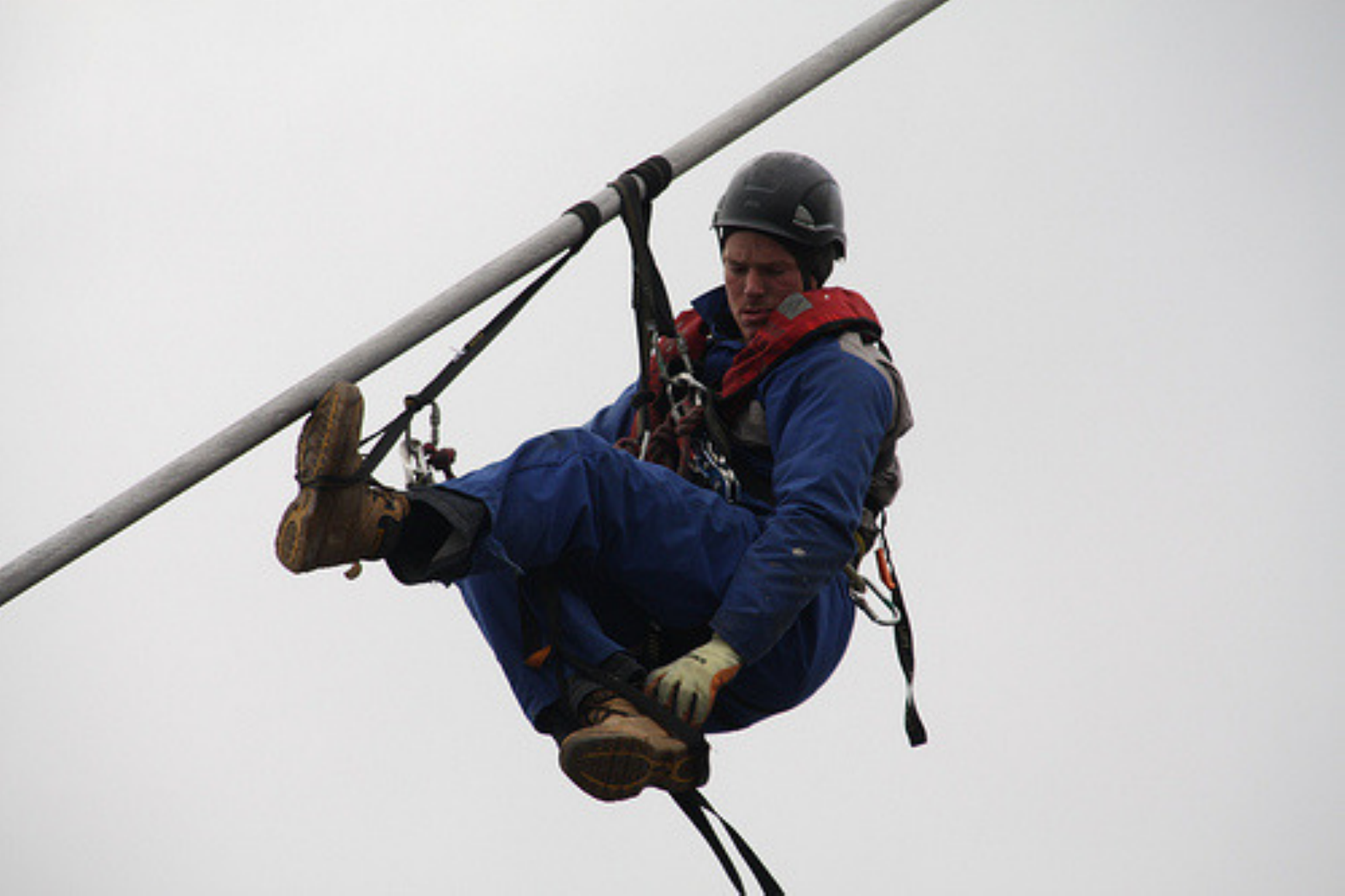}
    
    \raggedright
    (a) {\color{blue}a man in a blue shirt.}\\
    (b) {\color{red}a man in blue shirt.}\\
    
  \end{subfigure}
  \begin{subfigure}[t]{0.33\textwidth}
    \centering
    \includegraphics[width=3cm,height=2cm]{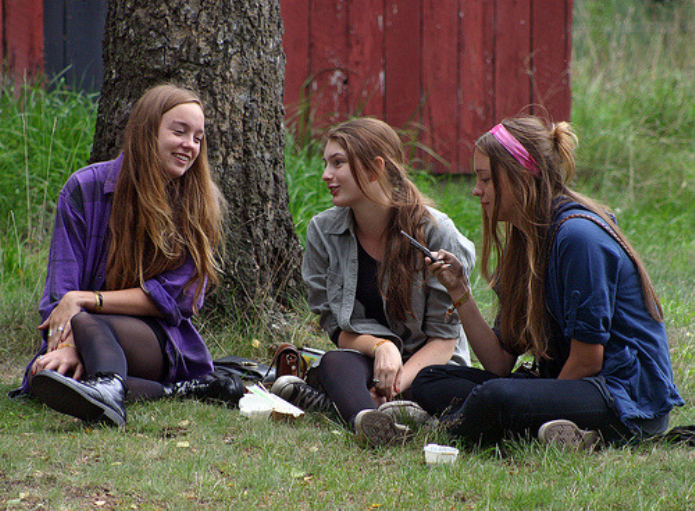}
    
    \raggedright
    (a) {\color{blue}three women are sitting.}\\
    (b) {\color{red}three women are sitting.}\\
    
  \end{subfigure}
  \begin{subfigure}[t]{0.3\textwidth}
    \centering
    \includegraphics[width=3cm,height=2cm]{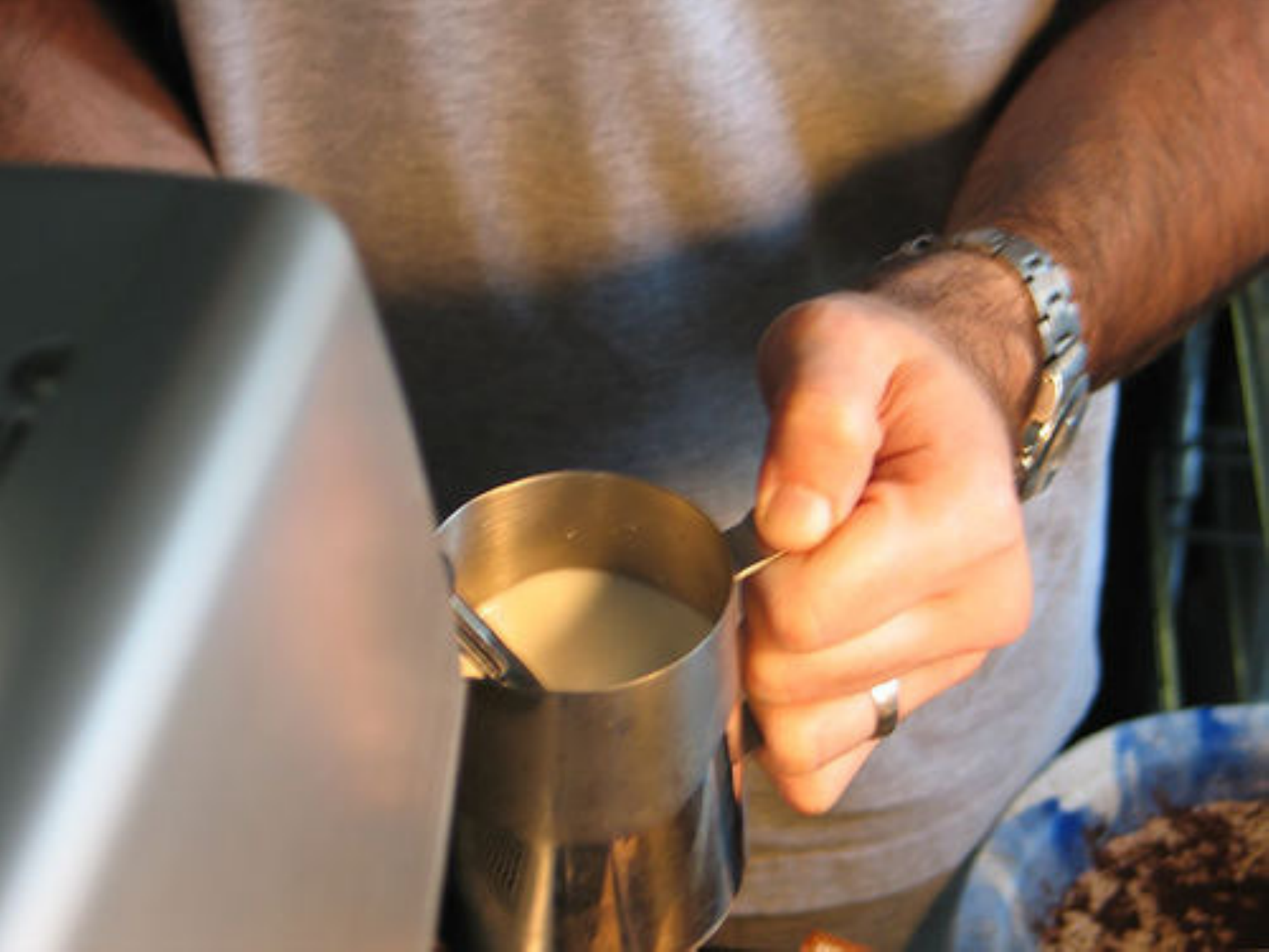}
    
    \raggedright
    (a) {\color{blue}a drink.}\\
    (b) {\color{red}a drink.}\\
    
  \end{subfigure}
  \vspace{10pt}
  
  \begin{subfigure}[t]{0.33\textwidth}
    \centering
    \includegraphics[width=3cm,height=2cm]{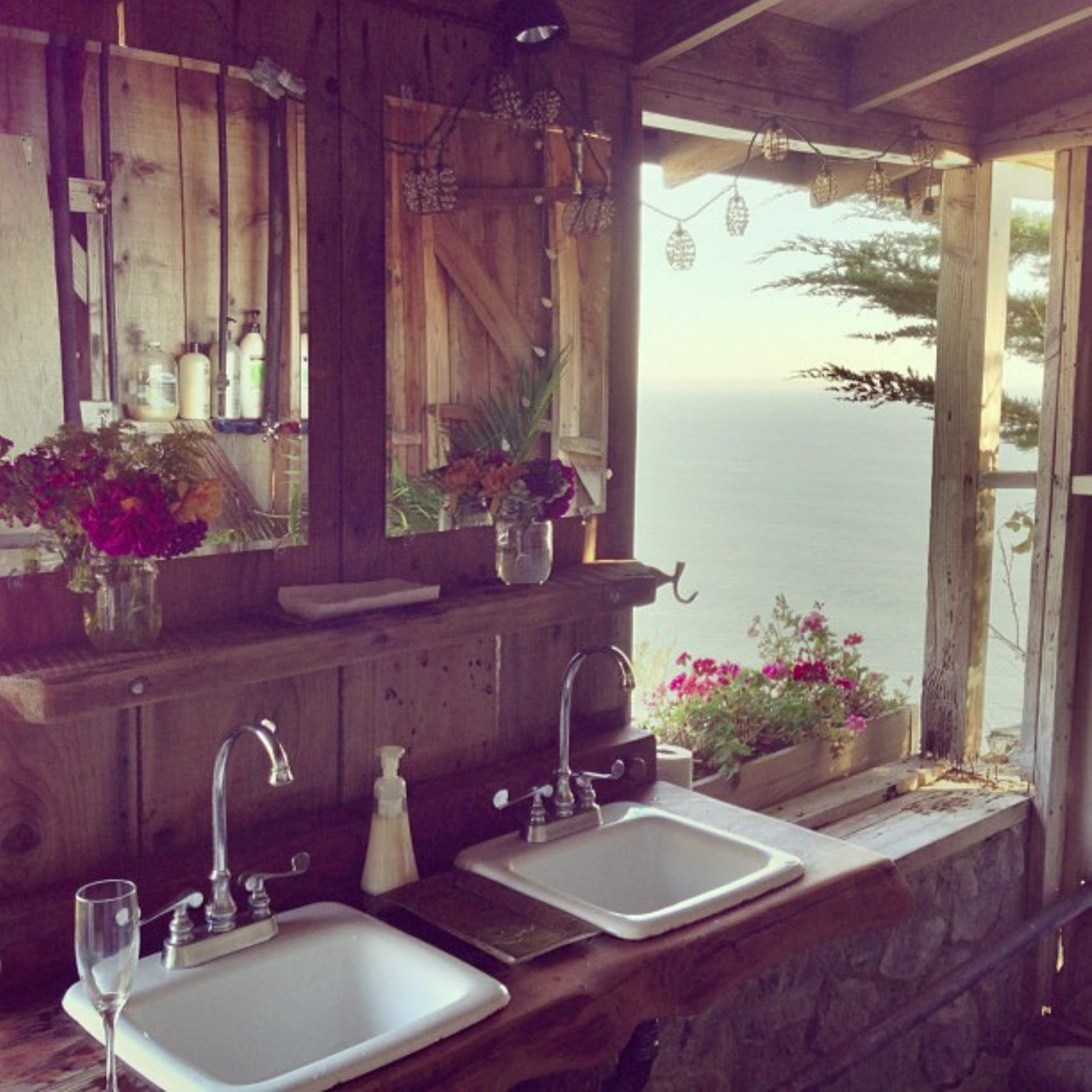}
    
    \raggedright
    (a) {\color{blue}a bathroom with a sink and a sink.}\\
    (b) {\color{red}a bathroom with a sink and a sink.}\\
    
  \end{subfigure}
  \begin{subfigure}[t]{0.33\textwidth}
    \centering
    \includegraphics[width=3cm,height=2cm]{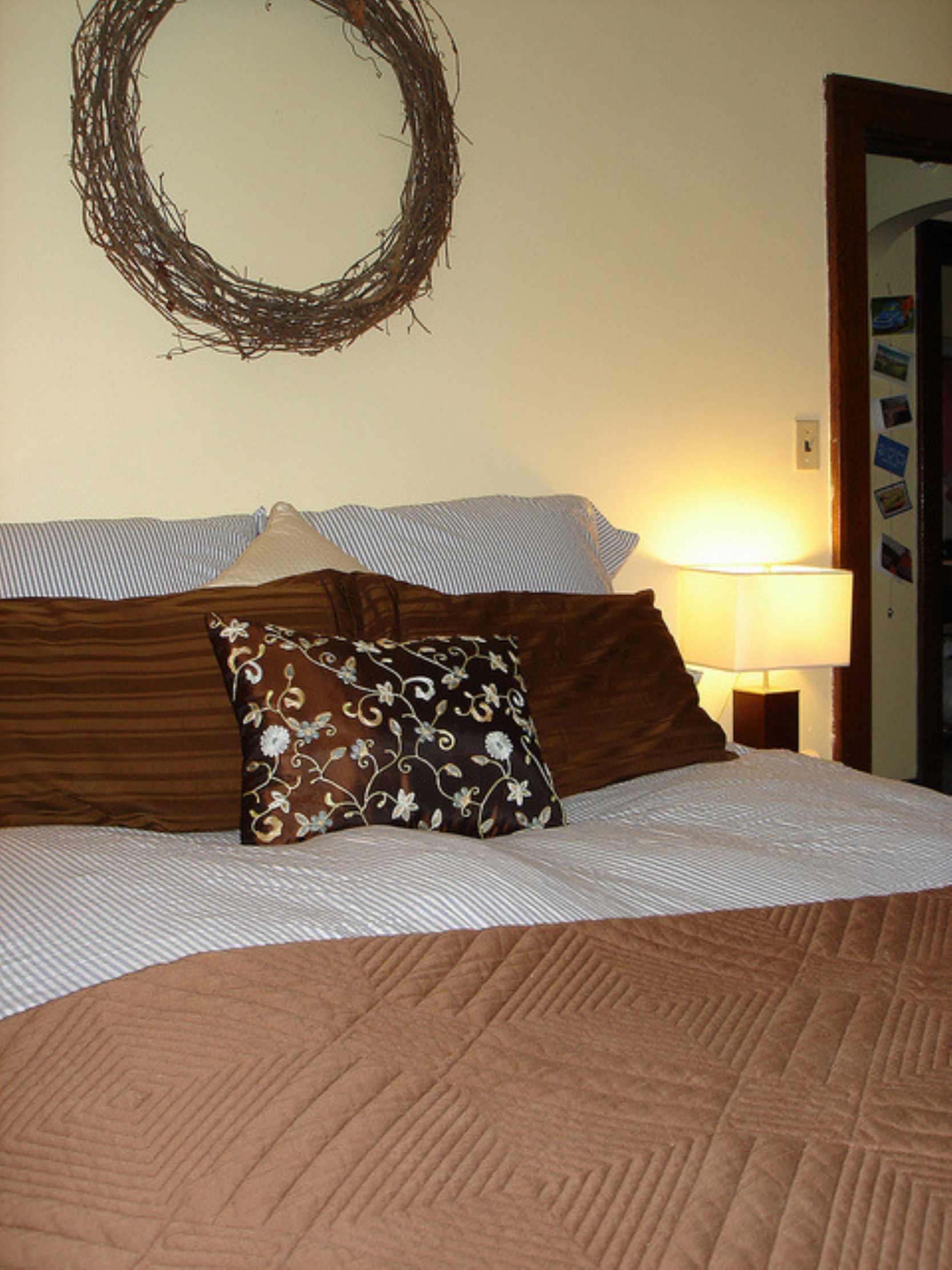}
    
    \raggedright
    (a) {\color{blue}a bed with a bed and a bed.}\\
    (b) {\color{red}a bed with a bed.}\\
    
  \end{subfigure}
  \begin{subfigure}[t]{0.3\textwidth}
    \centering
    \includegraphics[width=3cm,height=2cm]{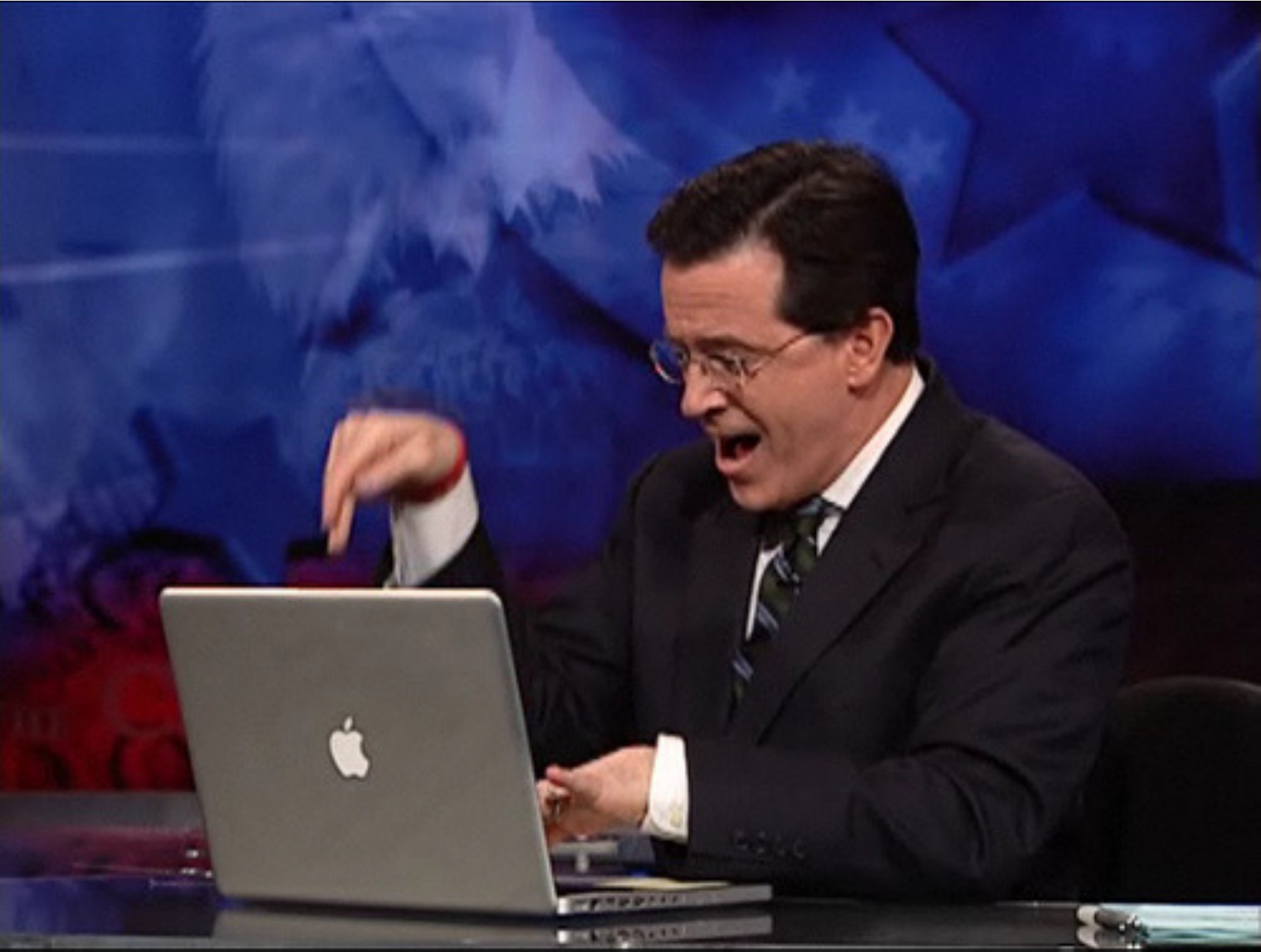}
    
    \raggedright
    (a) {\color{blue}a man sitting on top of a laptop computer.}\\
    (b) {\color{red}a man sitting on a laptop.}\\
    
  \end{subfigure}
  
  \caption{Comparison of captions generated by (a) Passport \cite{fan2019rethinking} with correct passport, (b) Passport \cite{fan2019rethinking} with forged passport. The first row is the images from Flickr30k. The second row is the images from the MS-COCO dataset.}
  \label{fig:passport-fake}
\end{figure}

\begin{table}[t]
\centering
\resizebox{0.95\textwidth}{!}{ 
\renewcommand*{\arraystretch}{1.5}
\Huge
\begin{tabular}{|c|cccccccc|cccccccc|}
\hline
\multirow{2}{*}{Methods} & \multicolumn{8}{c|}{MS-COCO} & \multicolumn{8}{c|}{Flickr30k}\\ \cline{2-17} 
  & B-1 & B-2 & B-3 & B-4 & M & R & C & S & B-1 & B-2 & B-3 & B-4 & M & R & C & S \\ \hline
Passport & 68.50 & 53.30 & 38.41 & 29.12 & 21.03 & 48.80 & 84.45 & 15.32 & 48.30 & 38.23 & 26.21 & 17.88 & 15.02 & 32.25 & 28.22 & 9.98 \\ 
$\bar{Passport}$ (forged) & 67.50 & 52.65 & 37.15 & 29.01 & 20.95 & 47.90 & 83.00 & 15.00 & 47.30 & 37.87 & 26.01 & 17.10 & 14.82 & 31.88 & 26.50 & 9.90 \\ 
\hline
\end{tabular}
}
  \caption{Comparison between Passport \cite{fan2019rethinking} with (top) correct passport and (bottom) forged passport on MS-COCO and Flickr30k datasets, across 5 common metrics where B-N, M, R, C, and S are BLEU-N, METEOR, ROUGE-L, CIDEr-D, and SPICE scores. It can be clearly seen that Passport model \cite{fan2019rethinking} is clearly insufficient to protect the image captioning model as the performance with correct and incorrect passport across all 5 metrics are almost similar. }
  \label{tab:passport}
\end{table}

\subsection{Comparison with CNN-based watermarking framework}
\label{cnnwatermark}

For comparison with the existing digital watermarking framework, we re-implement \cite{fan2019rethinking} using the official repository and refer this model as Passport. We choose \cite{fan2019rethinking} because the work is somehow similar to ours in terms of technical implementation. In Table \ref{tab:baseline}, we can notice that the overall performance of the Passport model on both MS-COCO and Flickr30k is very poor compared with the baseline and our proposed methods. For example, the CIDEr-D score dropped 10.45\% (MS-COCO) and 32.49\% (Flickr30K), respectively when compared to the baseline. In contrast, both of our proposed methods only dropped 3-4\% on both MS-COCO and Flickr30K when compared to the baseline. 

In terms of qualitative comparison, Figure \ref{fig:qualitative-passport} shows the sample captions generated by the Passport model are relatively brief when compared to the baseline and our approaches. For instance, the first image in Figure \ref{fig:qualitative-passport}, our proposed model generated {\it a woman in a scarf is standing}, it matches with the ground truth provided by the baseline, but the Passport model only generated {\it a woman}, missing the rest of the rich context altogether. a similar observation is found for the rest of the images.

Furthermore, we conducted an experiment to attack the Passport model with a forged passport and found out the Passport model still can have a relatively high CIDEr-D score. In Table \ref{tab:passport}, we show the quantitative results of the Passport model with the correct passport vs. Passport model with a forged passport. We found out that the Passport model with forged passport still can achieve very similar results on MS-COCO and Flickr30k dataset as to the Passport model with correct passport. For example, it has a CIDEr-D score of 83.0 (84.45) and 26.5 (28.22) on both datasets. In terms of qualitative comparison, Figure \ref{fig:passport-fake} shows the captions generated by the Passport model with (a) correct and (b) forged passport. It can be noticed that both models generate almost similar captions in terms of word selection and caption length. As a conclusion, we deduce that the conventional digital watermarking framework is insufficient to protect the image captioning model.

\subsection{Fidelity Evaluation}
\label{fidelity}

Fidelity is defined as matching the performance of the original model. In this section, we show that our proposed embedding schemes do not degrade the overall model performance in terms of metrics, as well as the quality of the generated sentences. According to Table \ref{tab:baseline}, it shows that the overall performance of our approaches and baseline model on MS-COCO and Flickr30k in all 5 image captioning metrics. Specifically, we can observe that $M_\oplus$ performed the best as it out-performed baseline in BLEU1-3 score on MS-COCO dataset, and BLEU-1 in Flickr30K dataset, respectively. For the rest of the metric score, we can also observe that $M_{\oplus}$ came as 2nd best score. In contrast, \cite{fan2019rethinking} performed poorly with at least a 10\% drop in all metrics. 

Subsequently, Table \ref{tab:unique} shows the comparison of the uniqueness of generated caption from our approaches and baseline model. A caption is considered unique if the generated caption does not exist in the training dataset. On both datasets, it shows that our approaches have very similar uniqueness and average caption length compared to baseline. This is consistent with the caption generated shown in Figure \ref{fig:qualitative-passport}. For example, both of our models have an exact caption generated as to baseline on the first image. And subsequently, in the rest of the images, the choice of words generated (i.e., shirt, room, road, swinging) are also very consistent with the baseline.



\begin{figure*}
  \begin{subfigure}[t]{0.4\textwidth}
    \centering
    \resizebox{0.99\textwidth}{!}{ 
    \begin{tikzpicture}
    \begin{axis}[
        title={},
        width=6cm,height=5cm,
        xlabel={Dissimilarity between real and fake key (\%)},
        ylabel={CIDEr-D score (\%)},
        xmin=0, xmax=100,
        ymin=0, ymax=45,
        legend pos=north east,
        ymajorgrids=true,
        grid style=dashed,
        legend columns=2, 
    ]
    
    \addplot[color=red,mark=,]
        coordinates {
        (0,40.07)(10,36.5)(20,27.07)(30,19.4)(40,14.07)(50,7.57)(60,5.27)(70,5.53)(80,9.27)(90,5.8)(100,3.57)
        };
        \addlegendentry{$M_{\oplus}$}
    \addplot[color=cyan,mark=,]
        coordinates {
        (0,40.17)(10,40.1)(20,37.17)(30,30.17)(40,17.53)(50,9.83)(60,7.37)(70,7.13)(80,9.67)(90,11.77)(100,12.4)
        };
        \addlegendentry{$M_{\otimes}$}
        
    \end{axis}
    \end{tikzpicture}
    }
    \\(a) Flickr30k
  \end{subfigure}
  \begin{subfigure}[t]{0.4\textwidth}
    \centering
    \resizebox{0.99\textwidth}{!}{ 
    \begin{tikzpicture}
    \begin{axis}[
        title={},
        width=6cm,height=5cm,
        xlabel={Dissimilarity between real and fake key (\%)},
        ylabel={CIDEr-D score (\%)},
        xmin=0, xmax=100,
        ymin=0, ymax=100,
        legend pos=north east,
        ymajorgrids=true,
        grid style=dashed,
        legend columns=2, 
    ]
    
    \addplot[color=red,mark=,]
        coordinates {
        (0,91.4)(10,36.5)(20,27.07)(30,19.4)(40,14.07)(50,7.57)(60,5.27)(70,5.53)(80,9.27)(90,5.8)(100,3.57)
        };
        \addlegendentry{$M_{\oplus}$}
    \addplot[color=cyan,mark=,]
        coordinates {
        (0,91.6)(10,89.1)(20,86.7)(30,74.77)(40,63.27)(50,42.93)(60,36.43)(70,28.17)(80,25.77)(90,27.13)(100,27.47)
        };
        \addlegendentry{$M_{\otimes}$}
        
    \end{axis}
    \end{tikzpicture}
    }
    \\(b) MS-COCO
  \end{subfigure}
  \begin{subfigure}[t]{0.4\textwidth}
    \centering
    \resizebox{0.99\textwidth}{!}{ 
    \begin{tikzpicture}
    \begin{axis}[
        title={},
        width=6.5cm,height=5cm,
        xlabel={Dissimilarity between real and fake signature (\%)},
        ylabel={CIDEr-D score (\%)},
        xmin=0, xmax=100,
        ymin=0, ymax=45,
        legend pos=north east,
        ymajorgrids=true,
        grid style=dashed,
        legend columns=2, 
    ]
    
    \addplot[color=red,mark=,]
        coordinates {
        (0,40.07)(10,39.57)(20,38.1)(30,37.37)(40,34.83)(50,30.67)(60,28.2)(70,25.07)(80,22.73)(90,20.27)(100,17.6)
        };
        \addlegendentry{$M_{\oplus}$}
    \addplot[color=cyan,mark=,]
        coordinates {
        (0,40.17)(10,39.6)(20,37.4)(30,29.57)(40,17.63)(50,11.03)(60,8.3)(70,8.2)(80,10.77)(90,13.37)(100,13.97)
        };
        \addlegendentry{$M_{\otimes}$}
        
    \end{axis}
    \end{tikzpicture}
    }
    \\(c) Flickr30k
  \end{subfigure}
  \begin{subfigure}[t]{0.4\textwidth}
    \centering
    \resizebox{0.99\textwidth}{!}{ 
    \begin{tikzpicture}
    \begin{axis}[
        title={},
        width=6.5cm,height=5cm,
        xlabel={Dissimilarity between real and fake signature (\%)},
        ylabel={CIDEr-D score (\%)},
        xmin=0, xmax=100,
        ymin=0, ymax=100,
        legend pos=north east,
        ymajorgrids=true,
        grid style=dashed,
        legend columns=2,
    ]
    
    \addplot[color=red,mark=,]
        coordinates {
        (0,91.4)(10,87.63)(20,84.13)(30,79)(40,69.03)(50,60.13)(60,52.33)(70,47.23)(80,45.1)(90,43.47)(100,42.2)
        };
        \addlegendentry{$M_{\oplus}$}
    \addplot[color=cyan,mark=,]
        coordinates {
        (0,91.6)(10,88.97)(20,86.83)(30,76.07)(40,66.3)(50,47.07)(60,43.87)(70,37.53)(80,34.5)(90,34.83)(100,36.53)
        };
        \addlegendentry{$M_{\otimes}$}
        
    \end{axis}
    \end{tikzpicture}
    }
    \\(d) MS-COCO
  \end{subfigure}
  \caption{CIDEr-D on Flickr30k and MS-COCO under ambiguity attack on (a-b) key; (c-d) signature.}
  \label{fig:amb-1}
\end{figure*}
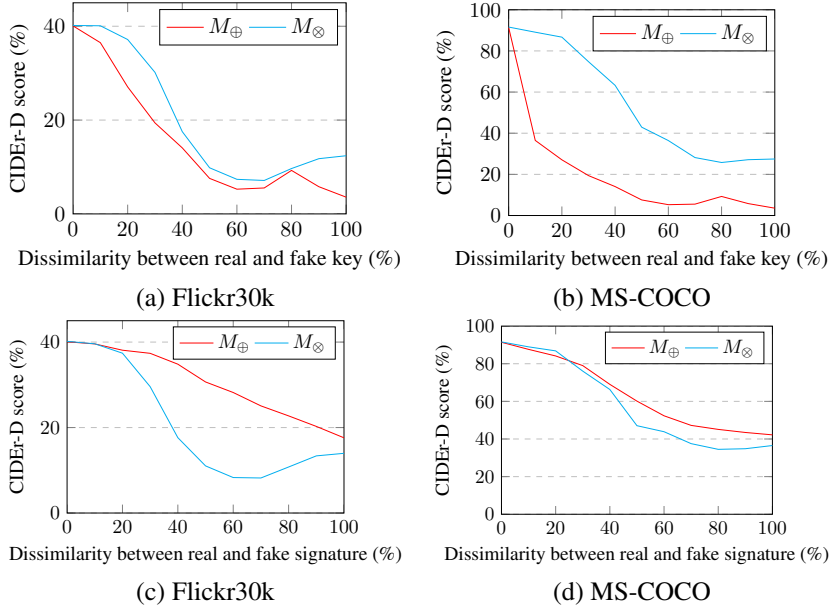

\begin{table}[t]
\centering
\resizebox{0.6\textwidth}{!}{ 
\renewcommand*{\arraystretch}{1.2}
\Huge
\begin{tabular}{|l|c|c|c|c|}
\hline
\multirow{2}{*}{Methods} & \multicolumn{2}{c|}{MS-COCO} & \multicolumn{2}{c|}{Flickr30k}\\ \cline{2-5} 
  & MS-COCO & Flickr30k & Flickr30k & MS-COCO \\ \hline
Baseline & - (94.30) & - (37.70) & - (41.80) & - (88.50) \\  
\hline
\(M_\oplus\) & 100 (91.40) &  70.40 (37.50) & 100 (40.07) & 72.50 (87.30)\\
\(M_\otimes\) & 99.99 (91.60) & 71.50 (37.8) & 99.99 (40.17) & 71.35 (86.50)\\
\hline
\end{tabular}
}
   \caption{Fine-tuning attack: CIDEr-D (in-bracket) of baseline and proposed models (Left: MS-COCO fine-tune on Flickr30k. Right: vice-versa.) Accuracy (\%) outside bracket is the signature detection rate.}
  \label{tab:removal-finetune}
\end{table}

\begin{table}[t]
\centering
\resizebox{0.6\textwidth}{!}{ 
\renewcommand*{\arraystretch}{1.1}
\begin{tabular}{|l|c|c|c|c|}
\hline
\multirow{2}{*}{Methods} & \multicolumn{2}{c|}{MS-COCO} & \multicolumn{2}{c|}{Flickr30k}\\ \cline{2-5} 
  & Unique & Avg-L& Unique & Avg-L \\ \hline
Baseline & 62.93\% & 8.86 & 88.80\% & 9.50 \\ 
\(M_\oplus\) & 70.96\% & 8.81 & 88.00\% & 9.30 \\ 
\(M_\otimes\) & 70.26\% & 8.91 & 87.10\% & 9.28 \\
\(\widehat{M_\oplus}\) & 100.00\% & 19.97 & 97.40\% & 18.56 \\ 
\(\widehat{M_\otimes}\) & 88.44\% & 12.71 & 53.40\% & 7.69 \\
\hline
\end{tabular}
}
   \caption{Comparison of the uniqueness of caption generated by our approaches and baseline model. \(\widehat{M_\oplus}\) and \(\widehat{M_\otimes}\) are \(M_\oplus\) and \(M_\otimes\), respectively but with forged secret key. Avg-L stands for average length.}
  \label{tab:unique}
\end{table}

\begin{figure*}[h]
\scriptsize
\centering
  \begin{subfigure}[t]{0.33\textwidth}
    \centering
    \includegraphics[width=3cm,height=2cm]{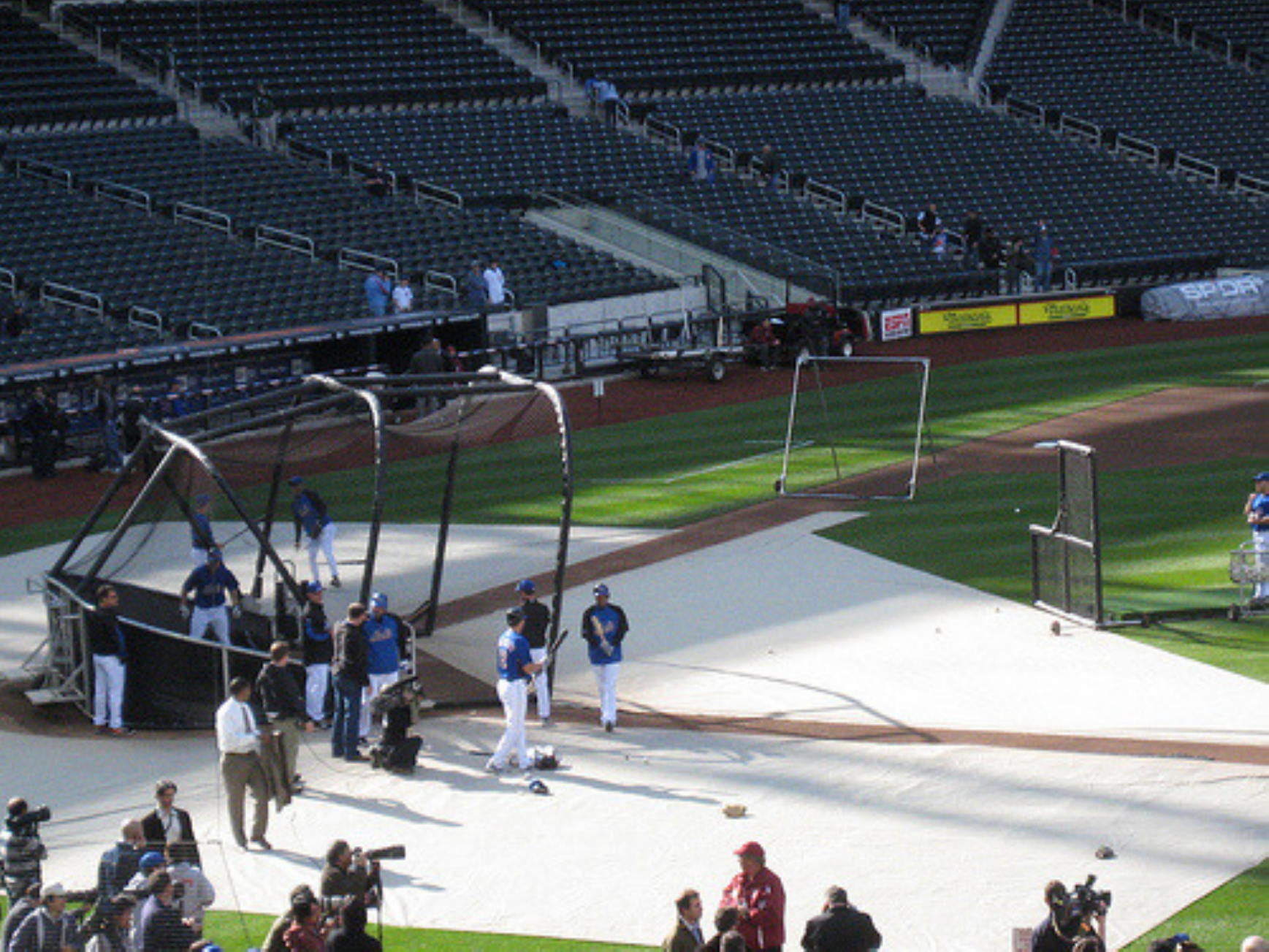}
    
    \raggedright
    \textbf{(a)} {\bf a baseball team is playing baseball.} \\
    (b) {\color{blue} a baseball team is playing baseball.}\\
    (c) a baseball game.\\
    (d) a baseball game.\\
  \end{subfigure}
  \begin{subfigure}[t]{0.33\textwidth}
    \centering
    \includegraphics[width=3cm,height=2cm]{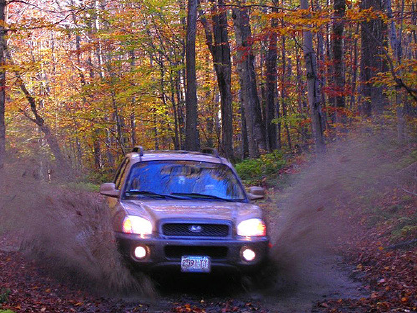}
    
    \raggedright
    \textbf{(a)} {\bf a car is driving through a puddle.}\\
    (b) {\color{blue} a dirt car is driving through the woods.}\\
    (c) {\color{red} a dirt car.}\\
    (d) {\color{red} a dirt car.}\\
  \end{subfigure}
  \begin{subfigure}[t]{0.32\textwidth}
    \centering
    \includegraphics[width=3cm,height=2cm]{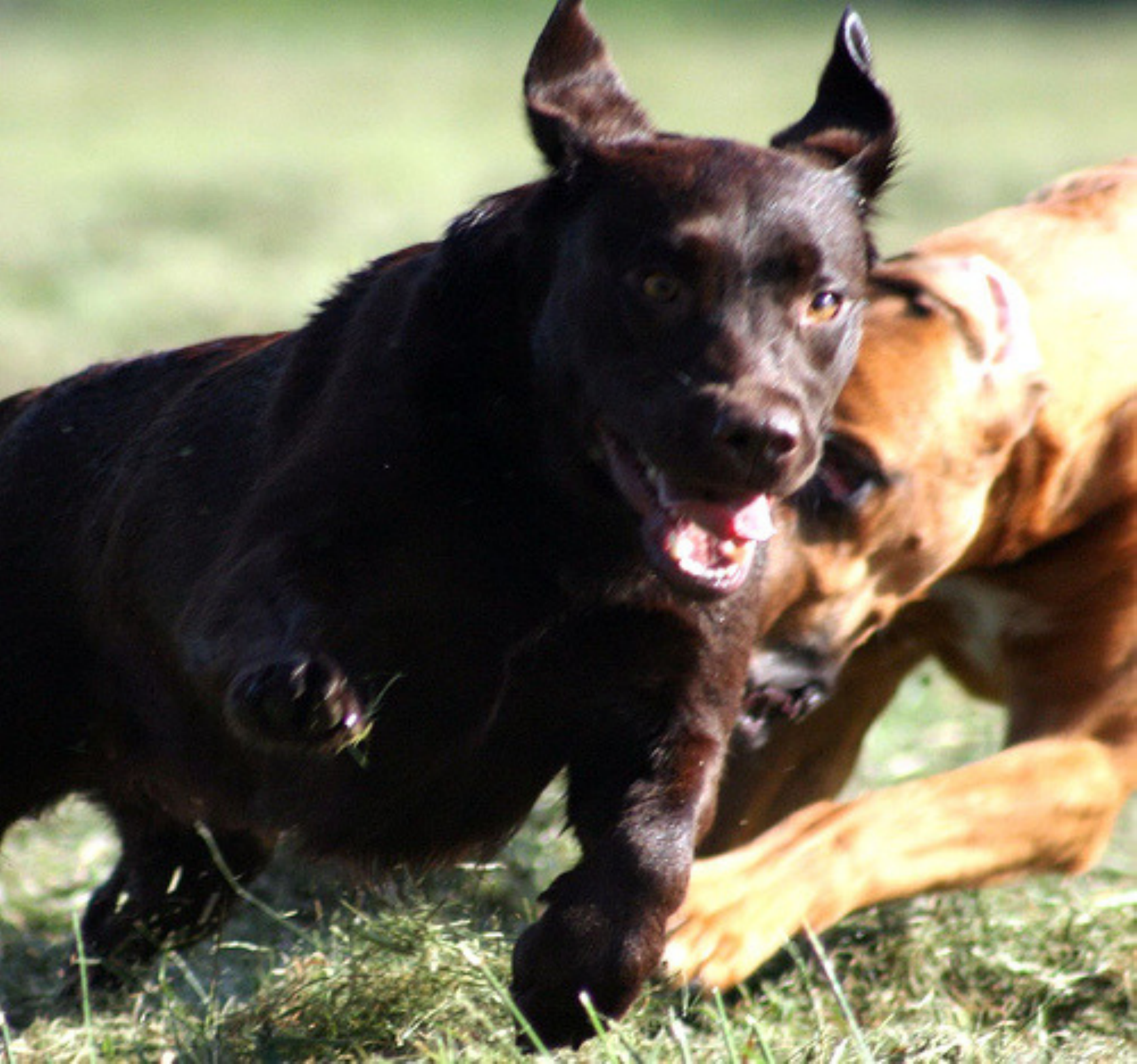}
    
    \raggedright
    \textbf{(a)} {\bf two dogs are running in the grass.}\\
    (b) {\color{blue} two dogs are running in the grass.}\\
    (c) two dogs are running.\\
    (d) {\color{red} two dogs.}\\
  \end{subfigure}
  \vspace{10pt}
  
  \begin{subfigure}[t]{0.33\textwidth}
    \centering
    \includegraphics[width=2cm,height=2cm]{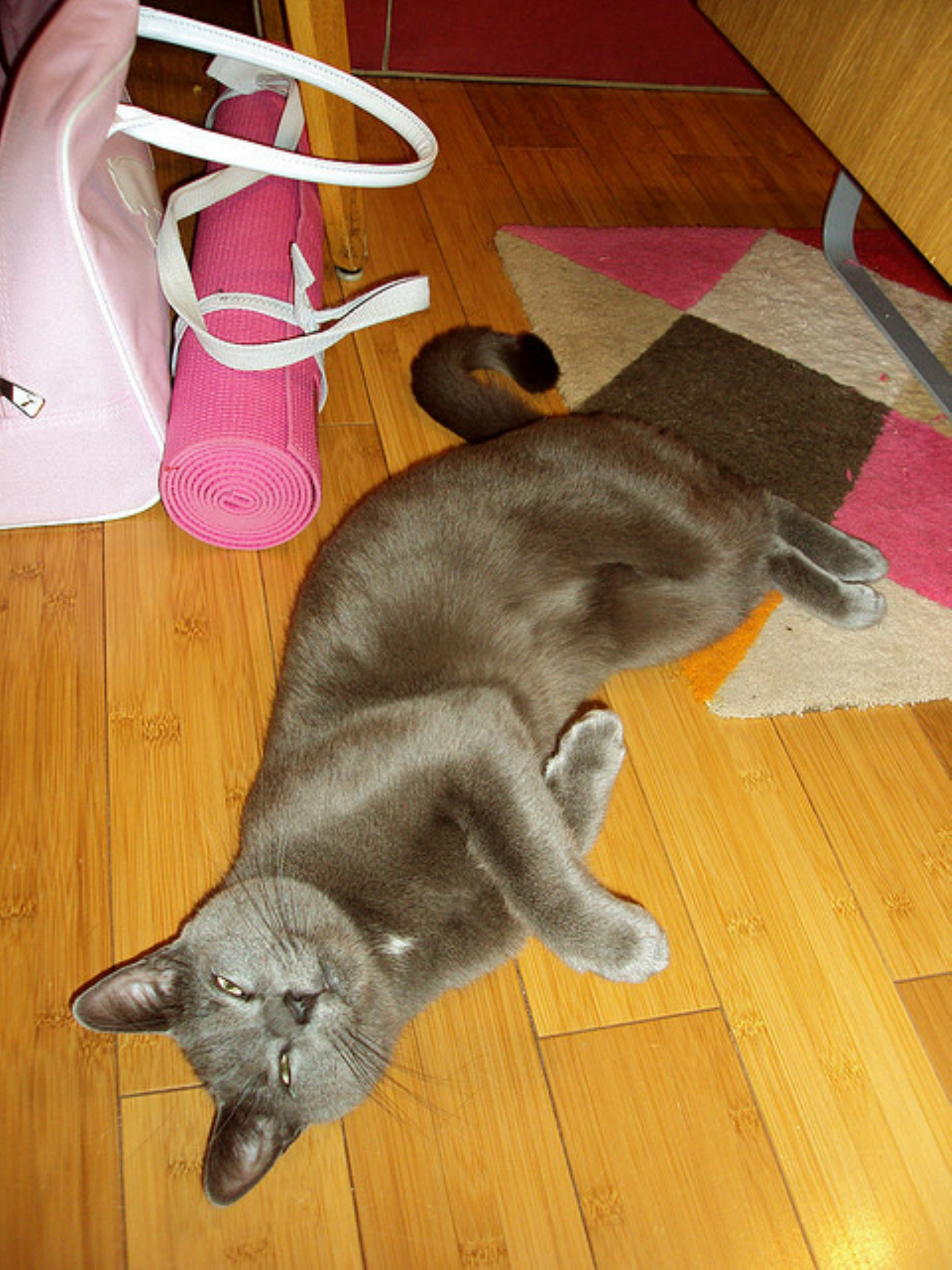}
    
    \raggedright
    \textbf{(a)} {\bf a cat laying on top of a wooden floor.}\\
    (b) {\color{blue} a cat laying on the floor of a wooden floor.}\\
    (c) {\color{red} a cat standing on a wooden floor.}\\
    (d) {\color{red} a cat on the floor on the floor on a wooden floor.}\\
  \end{subfigure}
  \begin{subfigure}[t]{0.33\textwidth}
    \centering
    \includegraphics[width=3cm,height=2cm]{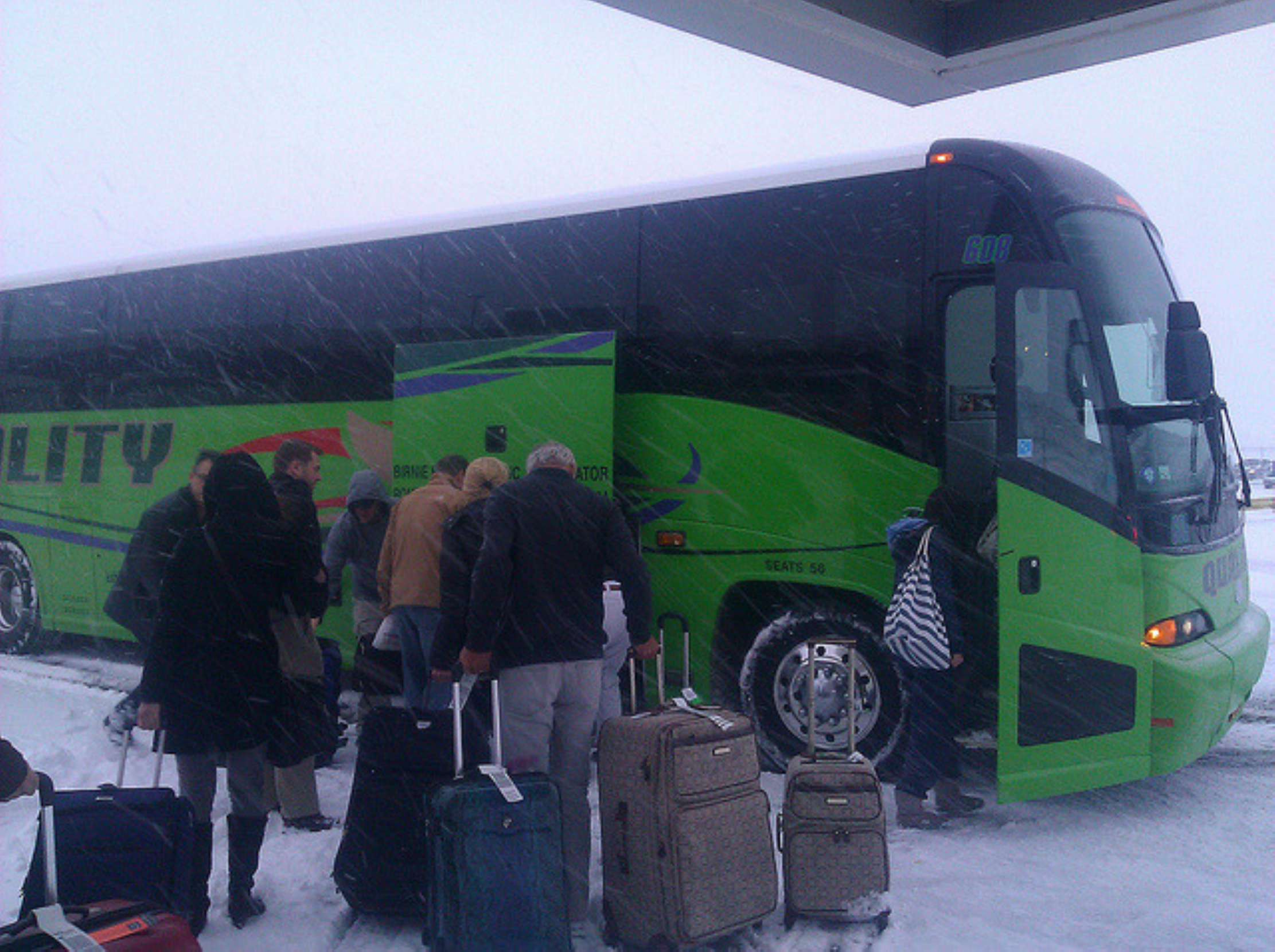}
    
    \raggedright
    \textbf{(a)} {\bf a group of people standing next to a truck.}\\
    (b) {\color{blue} a group of people standing next to a truck.}\\
    (c) {\color{red} a group.}\\
    (d) {\color{red} a group of people in a green and a large green and a large green and a large green and.}\\
  \end{subfigure}
  \begin{subfigure}[t]{0.32\textwidth}
    \centering
    \includegraphics[width=2cm,height=2cm]{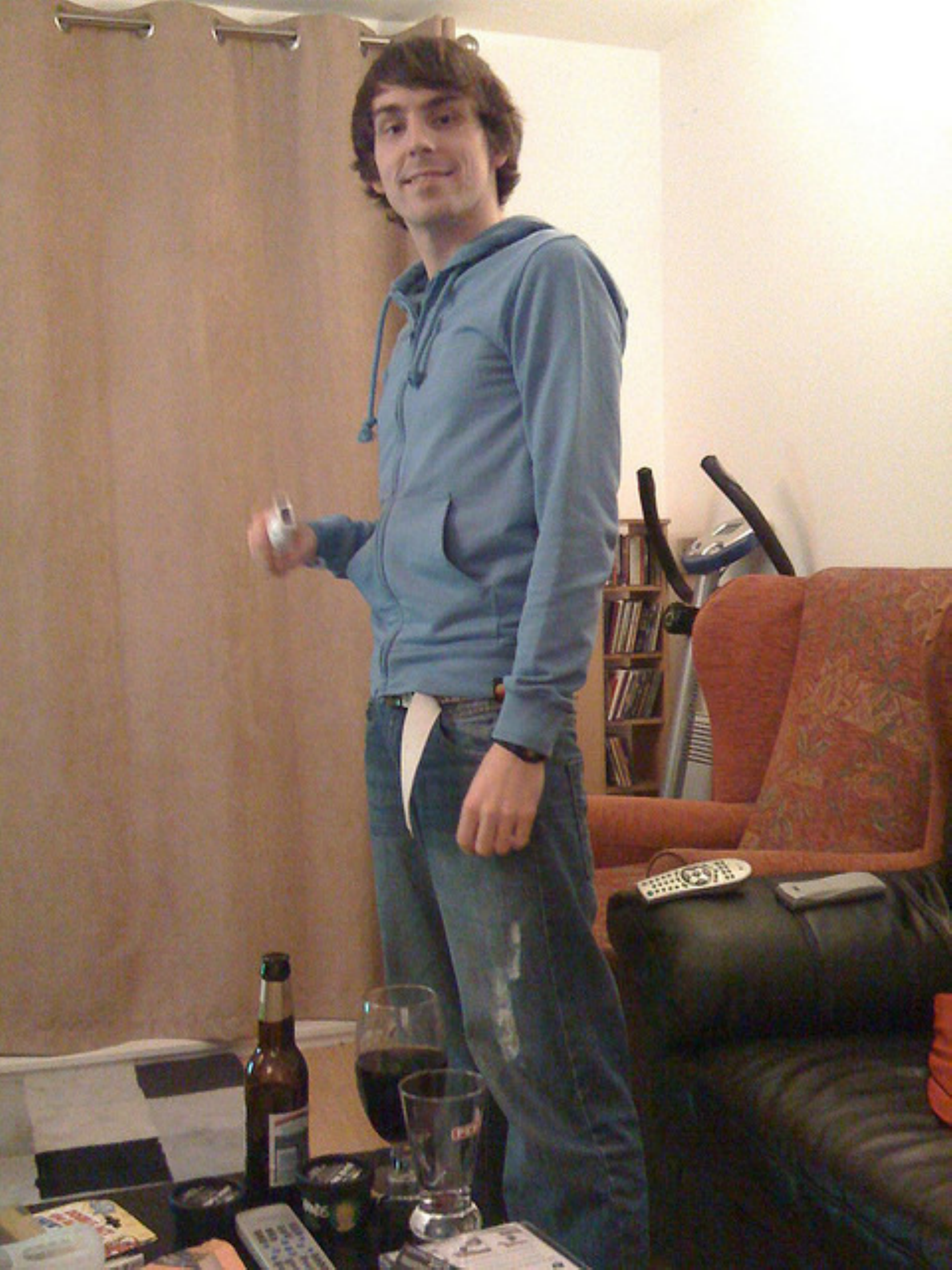}
    
    \raggedright
    \textbf{(a)} {\bf a man is standing in a living room.}\\
    (b) {\color{blue} a man is standing in a living room.}\\
    (c) {\color{red} a man and a man and a man and a man and a man and a man and a man.}\\
    (d) {\color{red} a man and a man in a man and a man in a man and a man and a man.}\\
  \end{subfigure}
  \caption{Comparison of captions generated from (a) Baseline, (b) $M_{\otimes}$, (c) $M_{\otimes}$ with the forged key has 75\% similarity as to real key and (d) $M_{\otimes}$ with the forged key has 50\% similarity as to real key. The first row is the images from Flickr30k. The second row is the images from the MS-COCO dataset.}
  \label{fig:qualitative}
  \vspace{-10pt}
\end{figure*}

\subsection{Resilience against ambiguity attacks}
\label{ambiguity}

\subsubsection{Protection against forged key}
In this case, we assume the attacker somehow has the access to the model but does not have the correct secret key and so tries to attack the model with a random forged key. Figure \ref{fig:amb-1}(a-b) show the CIDEr-D score of proposed models under ambiguity attack on the secret key in Flickr30k and MS-COCO.  Accordingly, we can observe that in general, the model performance will drop when a forged key is deployed. In particular, we would like to highlight that the CIDEr-D score on MS-COCO drops significantly (almost 50\% difference) in $M_{\otimes}$ even a forged key that has a 75\% similarity to the real key is being deployed. This shows that our proposed method is resilient against this forged key attack. 

Figure \ref{fig:qualitative} shows another six sample images and the respective captions generated by our proposal and baseline. From the first image, it shows that given the correct secret key, our proposed method is able to generate a caption that consists of object, scene, and attributes that are very similar to baseline. When a forged key is used, in this example, we show in Figure \ref{fig:qualitative}(c) a forged key that has a 75\% similarity to the correct secret key and in Figure \ref{fig:qualitative}(d) another forged key that has a 50\% matching to the correct secret key, the generated caption is either not meaningful at all with repetitive words (i.e., a man and a man) or a very brief caption (i.e., two dogs). According to Table \ref{tab:unique}, we can also observe similar patterns. For instance, $\widehat{M_\oplus}$ has almost 100\% uniqueness with the longest average caption length on both datasets. From Figure \ref{fig:qualitative}, we can understand that this is due to repetitive words. Meanwhile, $\widehat{M_\otimes}$ has 88.44\% uniqueness with the shortest average caption length on the Flick30K dataset. Yet again, this phenomenon is observed from the generated caption in Figure \ref{fig:qualitative}. 

\subsubsection{Protection against fake signature}
\begin{figure*}[h]
\scriptsize
\centering
  \begin{subfigure}[t]{0.33\textwidth}
    \centering
    \includegraphics[width=2.5cm,height=1.8cm]{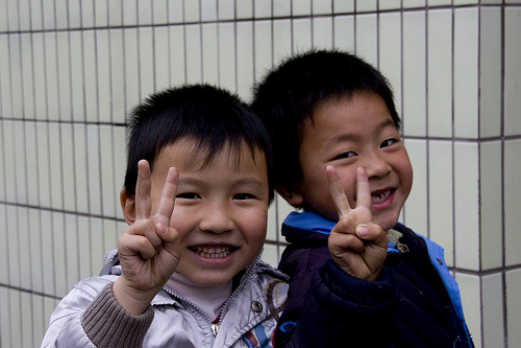}
    
    \raggedright
    \textbf{(a)} {\bf two young boys are posing for a sign.} \\
    (b) {\color{blue} two young boys are posing for a sign.}\\
    (c) two asian boys are holding a sign.\\
    (d) {\color{red} two asian asian asian asian asian asian boy in a sign in a sign.}\\
  \end{subfigure}
  \begin{subfigure}[t]{0.33\textwidth}
    \centering
    \includegraphics[width=2.5cm,height=1.8cm]{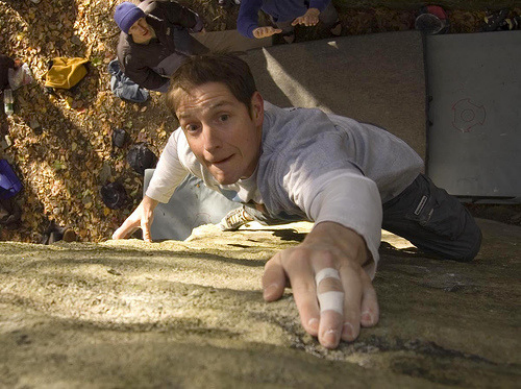}
    
    \raggedright
    \textbf{(a)} {\bf a man is climbing a rock.}\\
    (b) {\color{blue} a man is climbing a rock.}\\
    (c) a man is climbing.\\
    (d) {\color{red} a man in a rock climbing a rock rock climbing.}\\
  \end{subfigure}
  \begin{subfigure}[t]{0.32\textwidth}
    \centering
    \includegraphics[width=1.8cm,height=1.8cm]{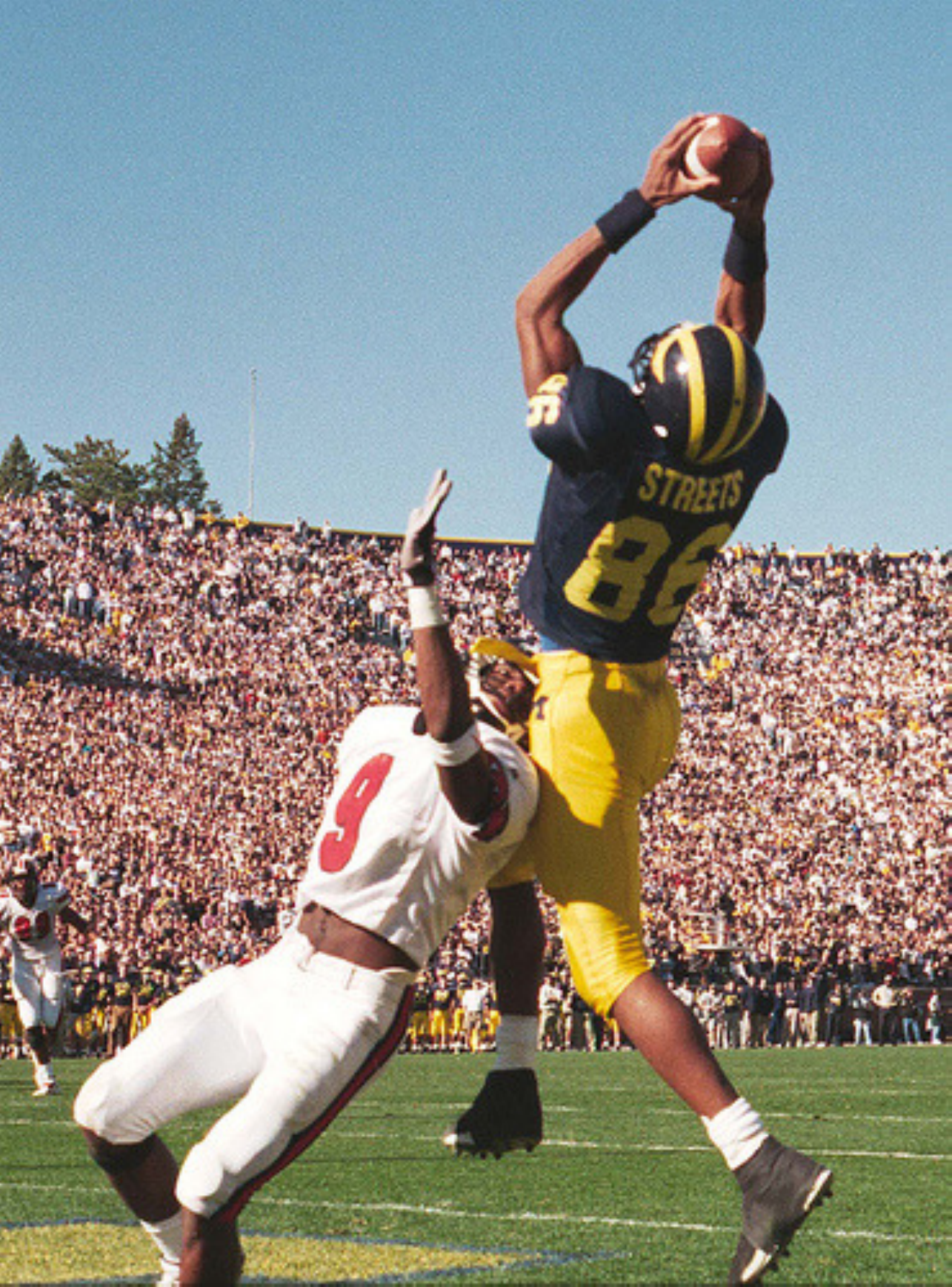}
    
    \raggedright
    \textbf{(a)} {\bf two football players are playing football in a field.}\\
    (b) {\color{blue} two football players are playing football.}\\
    (c) a football player playing football.\\
    (d) {\color{red} a football player in a football player in a football player in a football player in a football player in.}\\
  \end{subfigure}
  \begin{subfigure}[t]{0.32\textwidth}
    \centering
    \includegraphics[width=2.5cm,height=1.8cm]{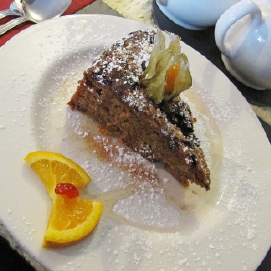}
    
    \raggedright
    \textbf{(a)} {\bf a piece of cake sitting on top of a plate.}\\
    (b) {\color{blue} a white plate topped with a piece of cake.}\\
    (c) a close up of a cake.\\
    (d) {\color{red} a white plate of a white plate with a white plate of a white plate of a white plate with.}\\
  \end{subfigure}
  \begin{subfigure}[t]{0.32\textwidth}
    \centering
    \includegraphics[width=2.5cm,height=1.8cm]{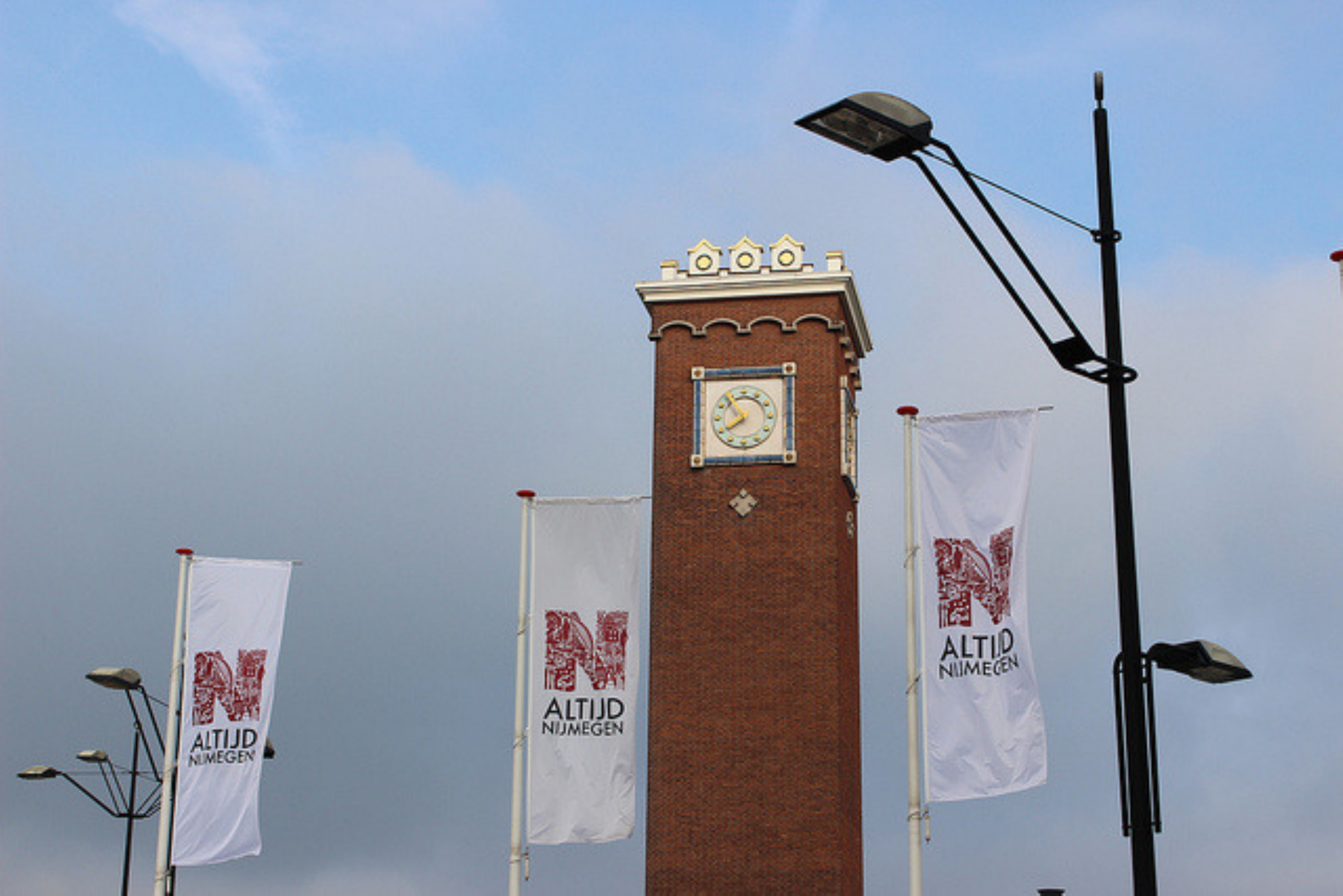}
    
    \raggedright
    \textbf{(a)} {\bf a building with a clock on top of it.}\\
    (b) {\color{blue} a building with a clock on top of it.}\\
    (c) {\color{red} a large building.}\\
    (d) {\color{red} a clock tower with a building with a building with a building with a building with a building with a.}\\
  \end{subfigure}
  \begin{subfigure}[t]{0.32\textwidth}
    \centering
    \includegraphics[width=2.5cm,height=1.8cm]{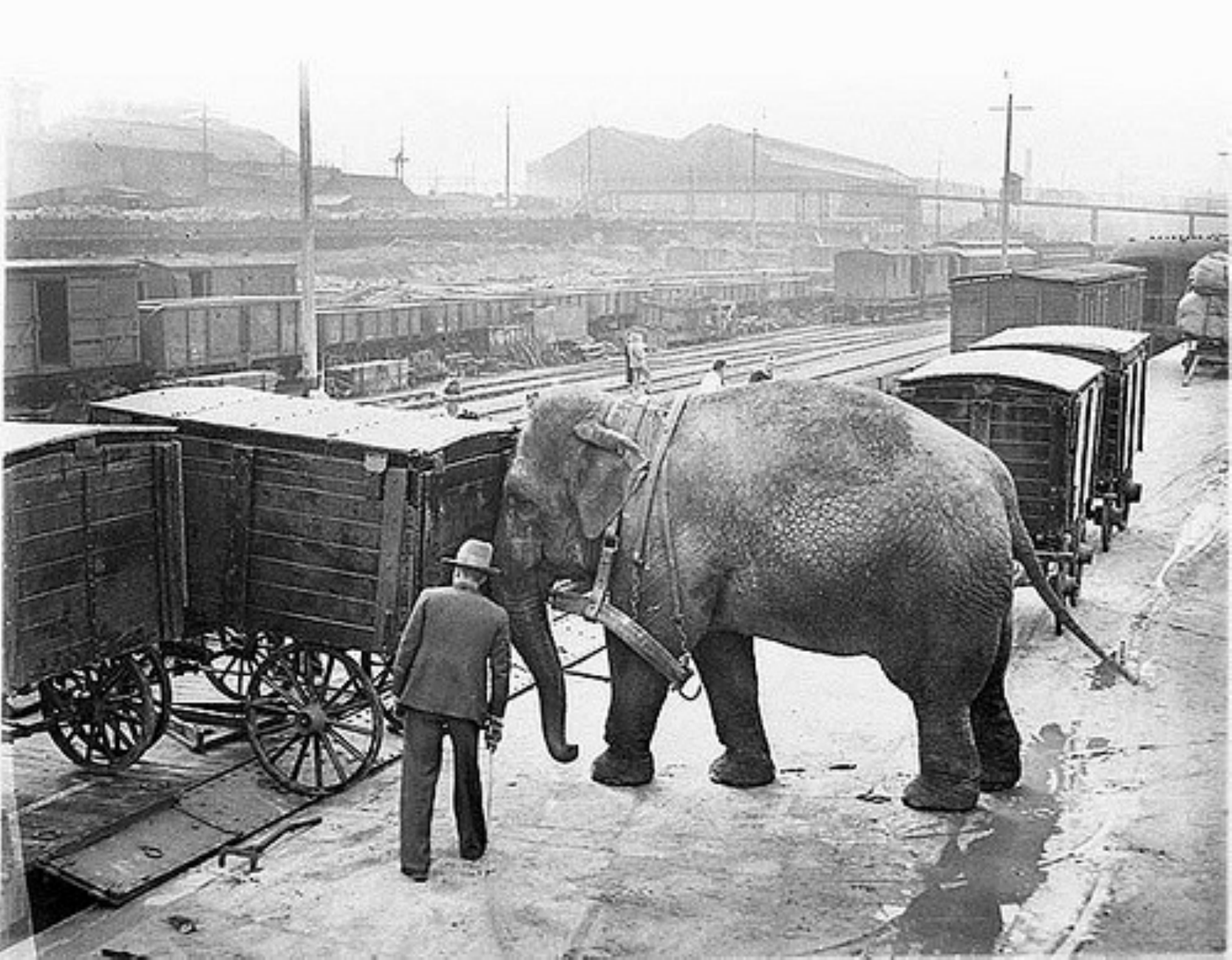}
    
    \raggedright
    \textbf{(a)} {\bf a man standing next to a small elephant.}\\
    (b) {\color{blue} a man standing next to an elephant next to an elephant.}\\
    (c) a man standing next to an elephant standing next to an elephant.\\
    (d) {\color{red} a man and a man and an elephant and a man and a man and a man and a man.}\\
  \end{subfigure}
  \caption{Comparison of captions generated from (a) Baseline, (b) $M_{\otimes}$, (c) $M_{\otimes}$ with 10\% of the sign are toggle and (d) $M_{\otimes}$ with 50\% of the sign are toggle. The first row is the images from Flickr30k.  The second row is the images from the MS-COCO dataset.}
  \label{fig:fakesign}
  \vspace{-13pt}
\end{figure*}

In this case, we assume the secret key is exposed to the attacker and one can use the model with original performance. However, the signature is able to use as proof of ownership. As such, the attacker will try to attack the signature by attempting to change the sign of the signature. Figure \ref{fig:amb-1}(c-d) show the overall performance of our proposed models (CIDEr-D score) will decrease when the signature is being compromised on both Flickr30k and MS-COCO. For instance, even very small changes (only 10\% of the sign are toggle), we can observe at least a 10-15\% drop of performance in terms of CIDEr-D score; and when half of the sign are toggle, it is seen that the model performance is almost useless. In Figure \ref{fig:fakesign}, when 10\% of the sign are modified, it can be seen that the captions generated by the model are relatively brief and shorter than the original model. For example in the second image, it shows that the generated caption is without ``a rock" when 10\% of the sign are modified. When 50\% of the sign are toggle, the captions are repetitive words. 

\subsection{Robustness against removal attacks}
\label{removal}

\subsubsection{Model pruning}

\begin{figure*}[t]
  \begin{subfigure}[t]{0.8\textwidth}
    \centering
    \resizebox{0.99\textwidth}{!}{ 
    \begin{tikzpicture}
    \begin{axis}[
        title={},
        width=6cm,height=5cm,
        xlabel={Pruning rate (\%)},
        ylabel={CIDEr-D score (\%)},
        xmin=0, xmax=100,
        ymin=0, ymax=45,
        legend pos=north east,
        ymajorgrids=true,
        grid style=dashed,
    ]
    
    \addplot[color=red,mark=,]
        coordinates {
        (0,40.07)(20,40.02)(40,38.93)(60,37.5)(80,30.5)(90,21.6)(100,4.1)
        };
        \addlegendentry{\(M_\oplus\)}
    \addplot[color=cyan,mark=,]
        coordinates {
        (0,40.17)(20,40.1)(40,39.4)(60,38.14)(80,31.9)(90,23.3)(100,5.84)
        };
        \addlegendentry{\(M_\otimes\)}
        
    \end{axis}
    \end{tikzpicture}
    \begin{tikzpicture}
    \begin{axis}[
        title={},
        width=6cm,height=5cm,
        xlabel={Pruning rate (\%)},
        ylabel={Signature detection rate (\%)},
        xmin=0, xmax=100,
        ymin=0, ymax=110,
        legend pos=north east,
        ymajorgrids=true,
        grid style=dashed,
    ]
    
    \addplot[color=red,mark=,]
        coordinates {
        (0,100)(20,100)(40,96)(60,85.4)(80,80.2)(90,62.25)(100,52)
        };
        \addlegendentry{\(M_\oplus\)}
    \addplot[color=cyan,mark=,]
        coordinates {
        (0,99.9)(20,99.9)(40,93)(60,84.5)(80,78.8)(90,60.5)(100,51.3)
        };
        \addlegendentry{\(M_\otimes\)}
        
    \end{axis}
    \end{tikzpicture}
    }
    \\(a) Flickr30k
  \end{subfigure}
  \begin{subfigure}[t]{0.8\textwidth}
    \centering
    \resizebox{0.99\textwidth}{!}{ 
    \begin{tikzpicture}
    \begin{axis}[
        title={},
        width=6cm,height=5cm,
        xlabel={Pruning rate (\%)},
        ylabel={CIDEr-D score (\%)},
        xmin=0, xmax=100,
        ymin=0, ymax=100,
        legend pos=north east,
        ymajorgrids=true,
        grid style=dashed,
    ]
    
    \addplot[color=red,mark=,]
        coordinates {
        (0,91.4)(20,90.2)(40,89)(60,87.6)(80,77.02)(90,60.8)(100,40.2)
        };
        \addlegendentry{\(M_\oplus\)}
    \addplot[color=cyan,mark=,]
        coordinates {
        (0,91.6)(20,91)(40,90.1)(60,89)(80,76.43)(90,61.12)(100,35.23)
        };
        \addlegendentry{\(M_\otimes\)}
        
    \end{axis}
    \end{tikzpicture}
    \begin{tikzpicture}
    \begin{axis}[
        title={},
        width=6cm,height=5cm,
        xlabel={Pruning rate (\%)},
        ylabel={Signature detection rate (\%)},
        xmin=0, xmax=100,
        ymin=0, ymax=110,
        legend pos=north east,
        ymajorgrids=true,
        grid style=dashed,
    ]
    
    \addplot[color=red,mark=,]
        coordinates {
        (0,100)(20,98)(40,94)(60,91.5)(80,80.5)(90,76.8)(100,51.8)
        };
        \addlegendentry{\(M_\oplus\)}
    \addplot[color=cyan,mark=,]
        coordinates {
        (0,99.99)(20,99.99)(40,96.5)(60,93.4)(80,88.5)(90,79.8)(100,50.45)
        };
        \addlegendentry{\(M_\otimes\)}
        
    \end{axis}
    \end{tikzpicture}
    }
    \\(a) MS-COCO
  \end{subfigure}
  
  \caption{Removal attack (Pruning): CIDEr-D score and signature detection rate of our approaches on both MS-COCO and Flickr30k against different pruning rates.}
  \label{fig:removal-pruning}
\end{figure*}
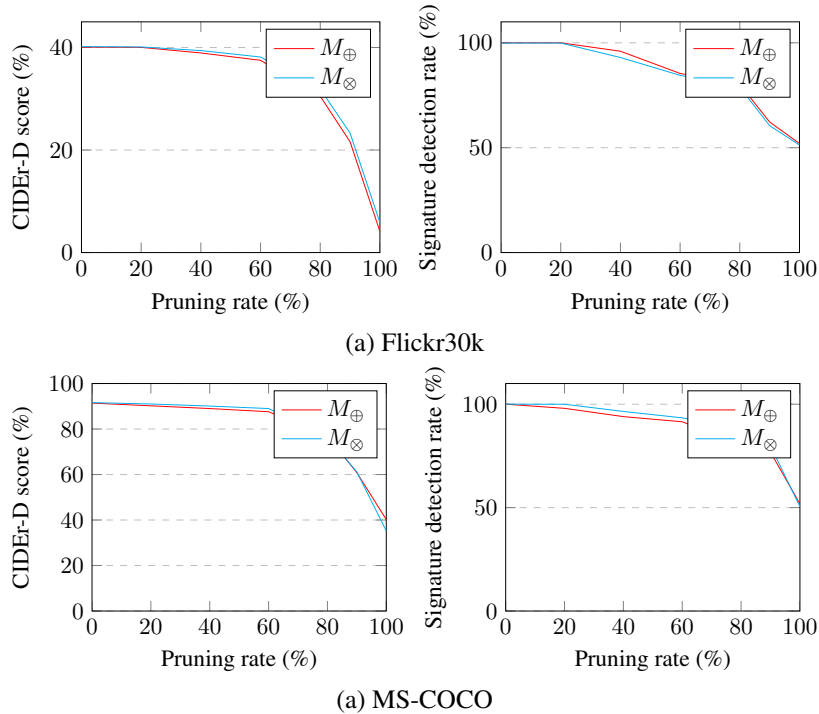

\begin{figure*}[h]
\scriptsize
\centering
  \begin{subfigure}[t]{0.3\textwidth}
    \centering
    \includegraphics[width=3cm,height=2cm]{supp/results/e_flickr_1-eps-converted-to.pdf}
    
    \raggedright
    \textbf{(a)} {\bf a dog jumps over a hurdle.} \\
    (b) {\color{blue} a dog jumps over a hurdle.}\\
    (c) a dog jumps over a hurdle.\\
  \end{subfigure}
  \begin{subfigure}[t]{0.33\textwidth}
    \centering
    \includegraphics[width=3cm,height=2cm]{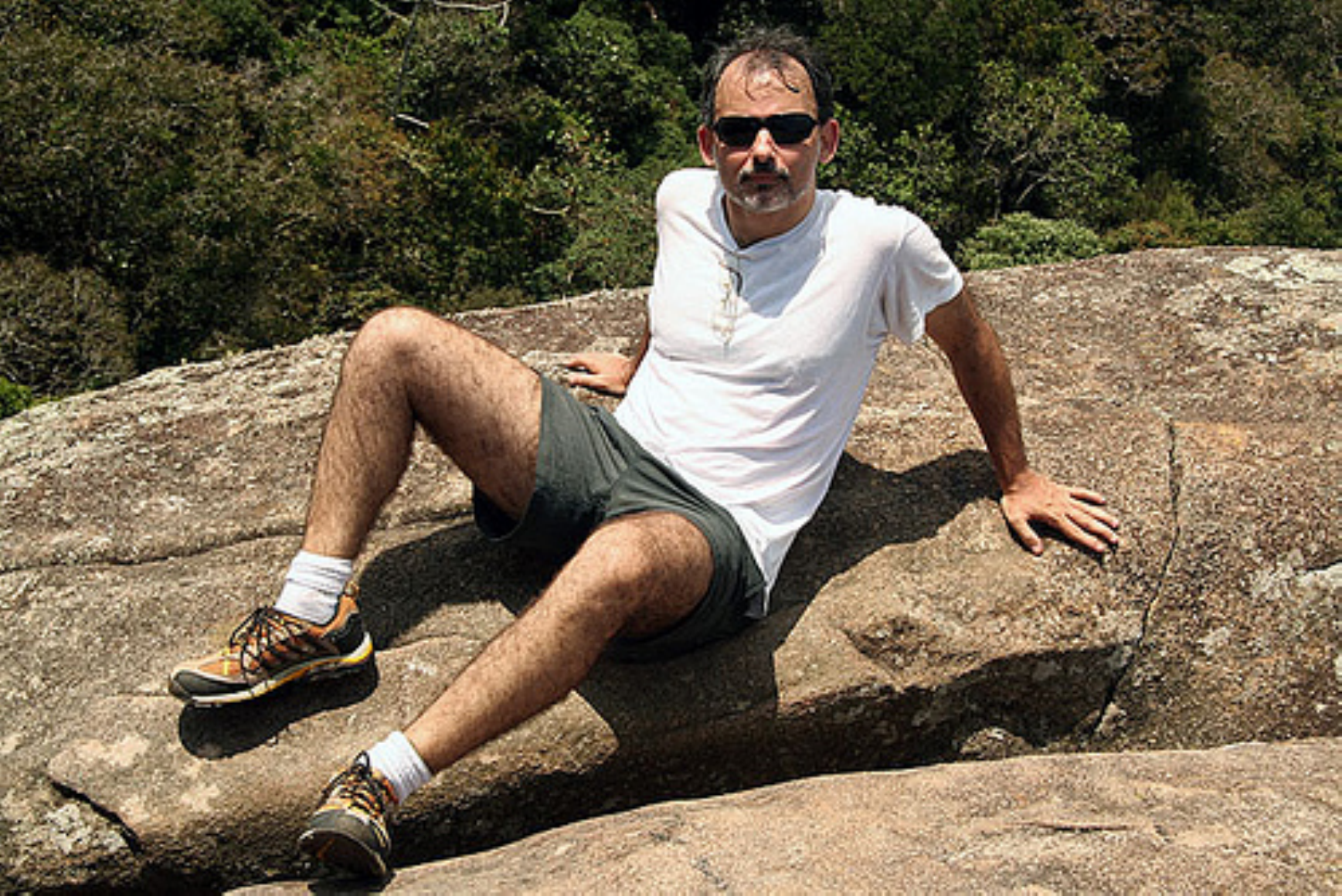}
    
    \raggedright
    \textbf{(a)} {\bf a man is sitting on a rock.} \\
    (b) {\color{blue} a man is sitting on a rock.}\\
    (c) a man sitting on rock.\\
  \end{subfigure}
  \begin{subfigure}[t]{0.33\textwidth}
    \centering
    \includegraphics[width=3cm,height=2cm]{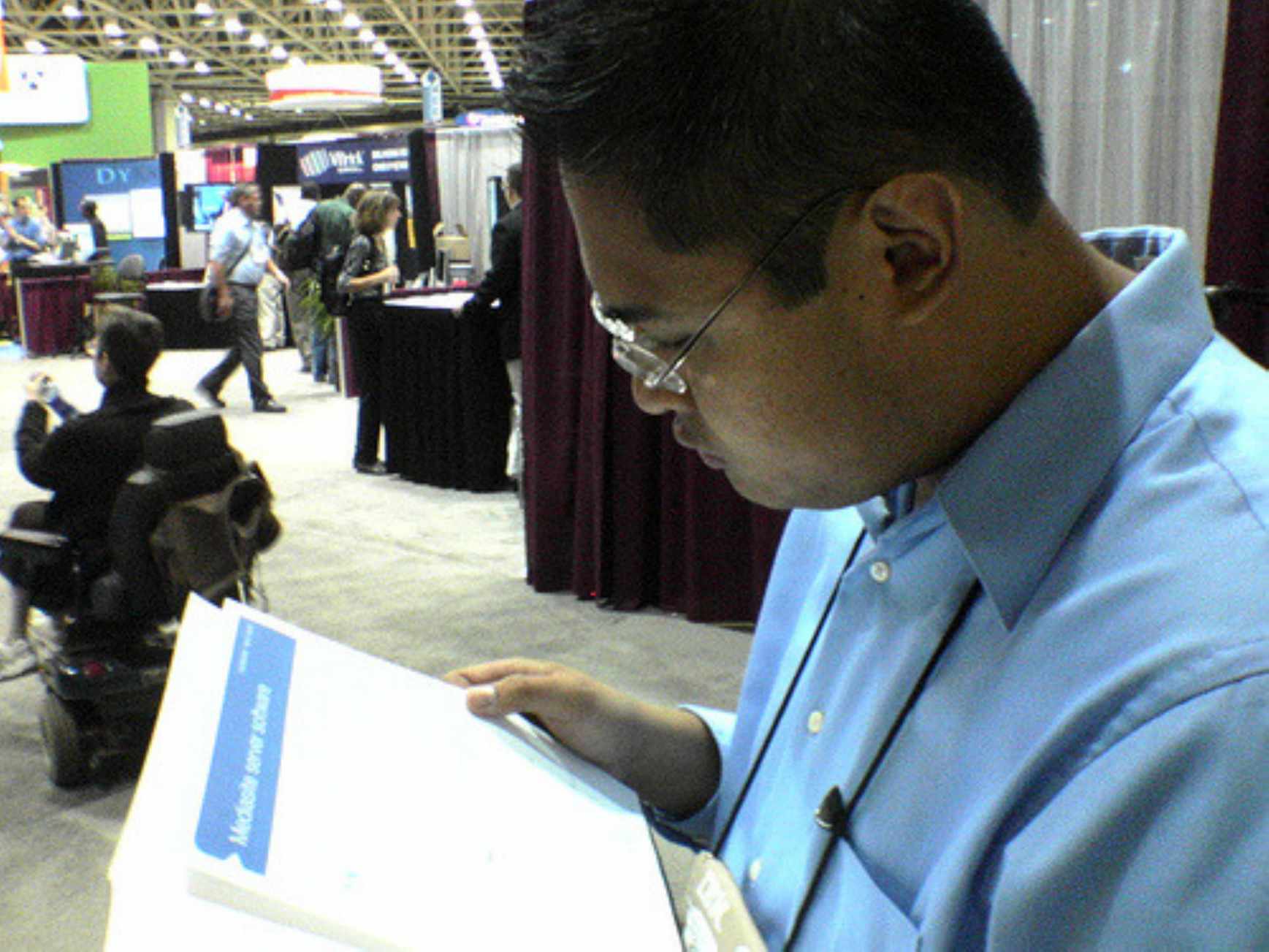}
    
    \raggedright
    \textbf{(a)} {\bf a man in a blue shirt is reading a book.} \\
    (b) {\color{blue} a man in a blue shirt is reading a book.}\\
    (c) a man is reading a book.\\
  \end{subfigure}
  
    \vspace{10pt}

  \begin{subfigure}[t]{0.33\textwidth}
    \centering
    \includegraphics[width=3cm,height=2cm]{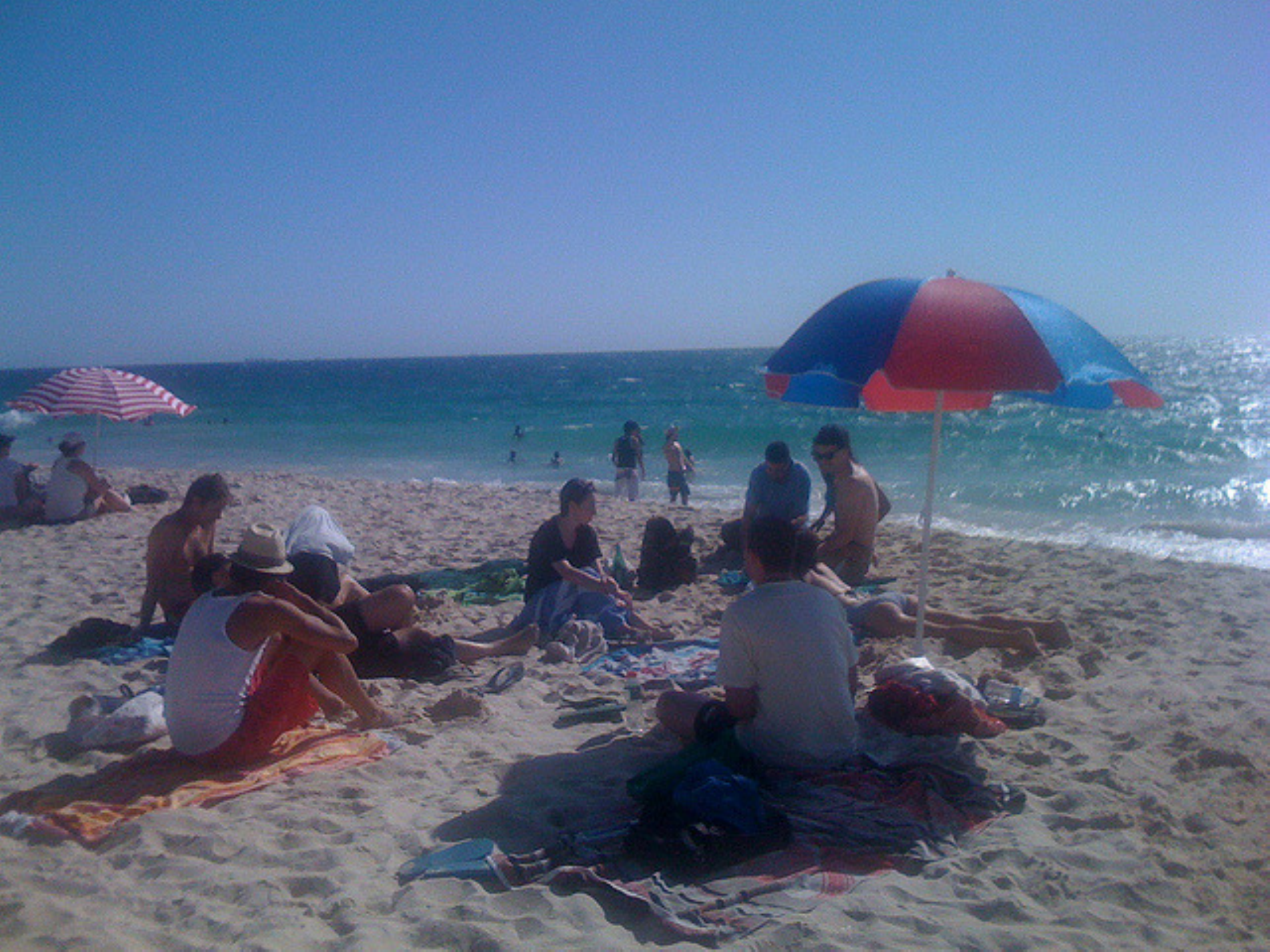}
    
    \raggedright
    \textbf{(a)} {\bf a group of people sitting on a beach.} \\
    (b) {\color{blue} a group of people sitting on a beach with umbrellas.}\\
    (c) a group of people sitting on a beach.\\
  \end{subfigure}
  \begin{subfigure}[t]{0.33\textwidth}
    \centering
    \includegraphics[width=3cm,height=2cm]{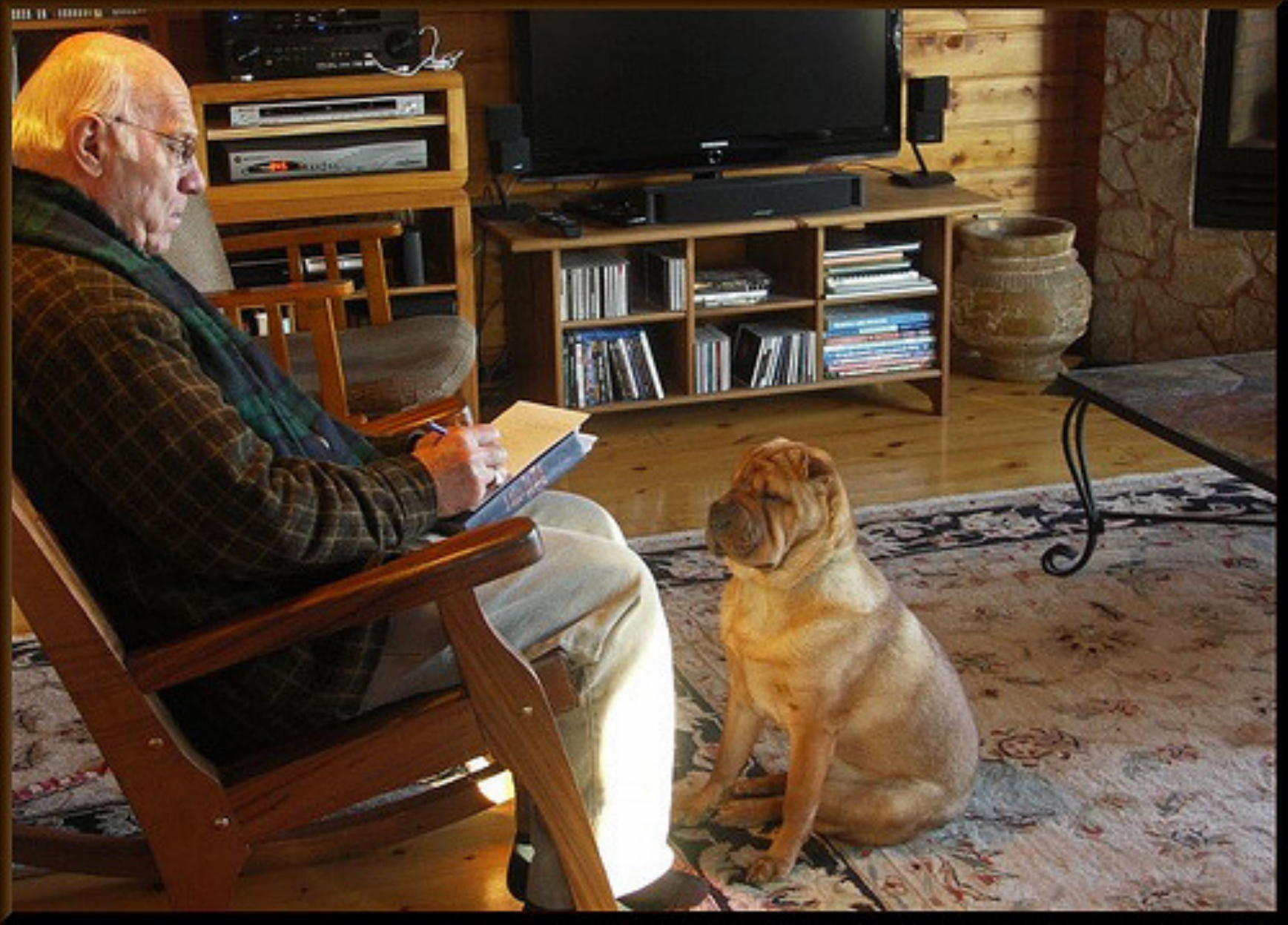}
    
    \raggedright
    \textbf{(a)} {\bf a dog sitting on a chair in front of a tv.} \\
    (b) {\color{blue} a man sitting on a chair next to a tv.}\\
    (c)a man is sitting on a chair in front of dog.\\
  \end{subfigure}
  \begin{subfigure}[t]{0.3\textwidth}
    \centering
    \includegraphics[width=3cm,height=2cm]{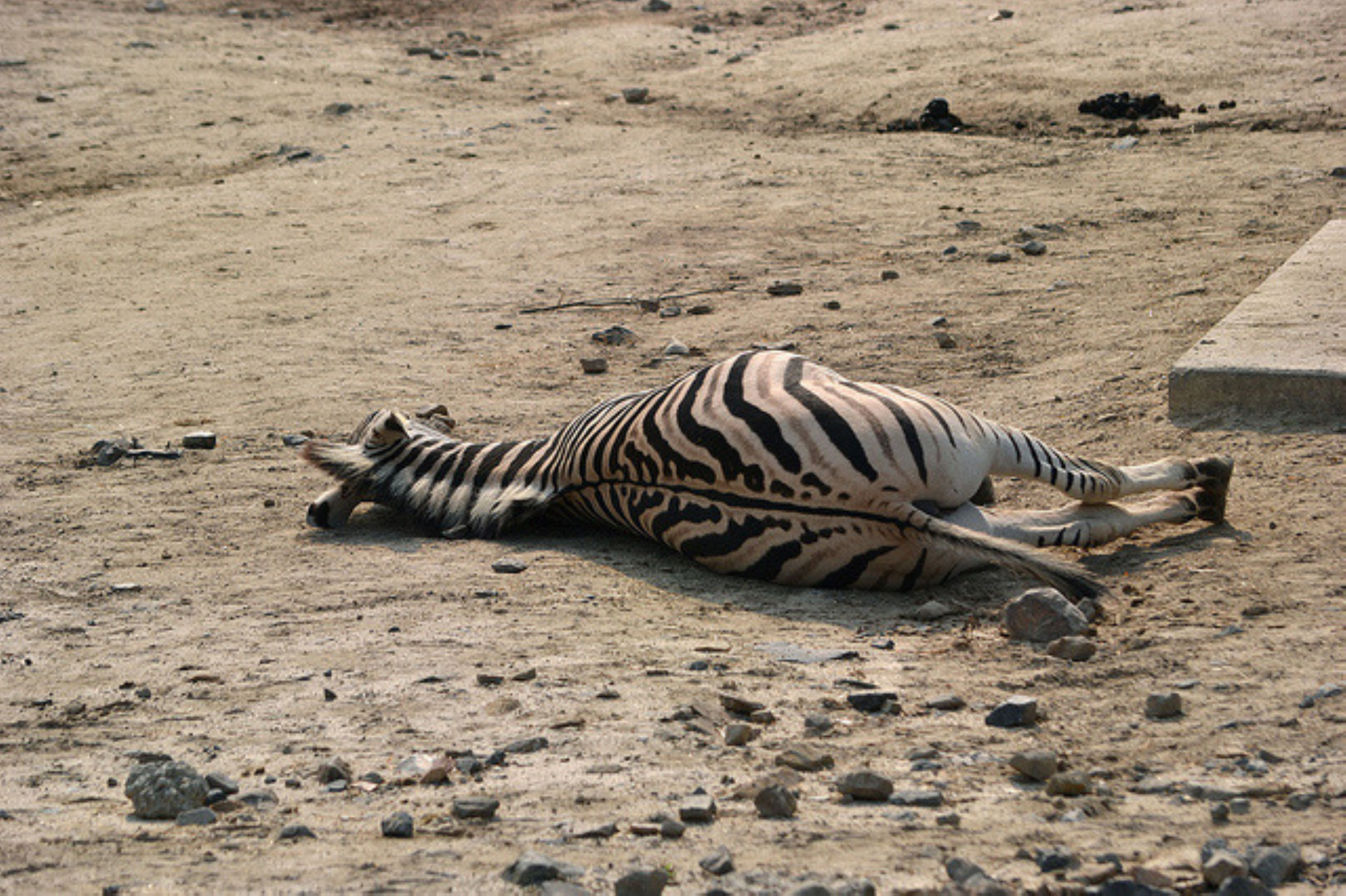}
    
    \raggedright
    \textbf{(a)} {\bf a zebra laying down on a sandy beach.} \\
    (b) {\color{blue} a zebra laying down on a dirt ground.}\\
    (c) a zebra laying down on ground.\\
  \end{subfigure}
  \caption{Comparison of captions generated from (a) Baseline, (b) $M_{\otimes}$, (c) $M_{\otimes}$ with 60\% pruning rate. The first row is the images from Flickr30k. The second row is the images from the MS-COCO dataset.}
  \label{fig:pruning}
\end{figure*}

Generally, model pruning is used to reduce the weights and computation overhead of a DNN model. However, the attacker might leverage it to remove the signature in the model. In order to test our approaches is robust to this attack, we implemented class-blind pruning method \cite{see2016compression}. Figure \ref{fig:removal-pruning} shows the CIDEr-D score and signature detection rate on \(M_\oplus\) and \(M_\otimes\) against different pruning rates.  We show that even 60\% of the network parameters are pruned, the signature detection rate is still intact at more than 84\% and 91\% on both Flickr30k and MS-COCO datasets, respectively. Figure \ref{fig:pruning} shows that the pruned model can still generate a meaningful caption for the images from both Flickr30k and MS-COCO datasets. As a summary, we show that even 60\% of the network parameters are pruned, the signature detection rate and the quality of the caption are still intact.

\subsubsection{Fine-tuning}
\label{subsubsec: fine-tuning}
Here, we simulate an attacker who fine-tunes the stolen model with a new dataset to obtain a new model that inherits the performance of the stolen model while attempting to remove the embedded signature. Table \ref{tab:removal-finetune} shows the signature detection rate and CIDEr-D score of the proposed model after perform fine-tuning. The signature can be detected at almost 100\% accuracy for our approaches in the original task. After fine-tuning the model (e.g., from MS-COCO to Flickr30k or Flickr30k to MS-COCO), we show that our approaches achieve comparable CIDEr-D score as to the baseline, however, we observe the signature detection rate decreased to around 70\%. This is one of the limitations of the proposed method but overall it does not compromise the IP protection of the model as we still have the secret key to act as proof of ownership. Therefore, the proposed secret key working together with the signature in this paper can act as complete protection for ownership verification.

\subsubsection{Key pruning}
\begin{figure*}[t]
  \begin{subfigure}[t]{0.8\textwidth}
    \centering
    \resizebox{0.99\textwidth}{!}{ 
    \begin{tikzpicture}
    \begin{axis}[
        title={},
        width=6cm,height=5cm,
        xlabel={Key pruning rate (\%)},
        ylabel={CIDEr-D score (\%)},
        xmin=0, xmax=100,
        ymin=0, ymax=45,
        legend pos=north east,
        ymajorgrids=true,
        grid style=dashed,
    ]
    
    \addplot[color=red,mark=,]
        coordinates {
        (0,40.07)(20,35.4)(40,27.5)(60,19.6)(80,10.2)(90,8.2)(100,6.5)
        };
        \addlegendentry{\(M_\oplus\)}
    \addplot[color=cyan,mark=,]
        coordinates {
        (0,40.17)(20,38.1)(40,35.3)(60,31.6)(80,22.3)(90,18.2)(100,15.1)
        };
        \addlegendentry{\(M_\otimes\)}
        
    \end{axis}
    \end{tikzpicture}
    \begin{tikzpicture}
    \begin{axis}[
        title={},
        width=6cm,height=5cm,
        xlabel={Key pruning rate (\%)},
        ylabel={Signature detection rate (\%)},
        xmin=0, xmax=100,
        ymin=0, ymax=110,
        legend pos=north east,
        ymajorgrids=true,
        grid style=dashed,
    ]
    
    \addplot[color=red,mark=,]
        coordinates {
        (0,100)(20,100)(40,99.99)(60,99.95)(80,99.41)(90,98.29)(100,95.54)
        };
        \addlegendentry{\(M_\oplus\)}
    \addplot[color=cyan,mark=,]
        coordinates {
        (0,99.9)(20,100)(40,100)(60,99.99)(80,99.95)(90,99.93)(100,99.79)
        };
        \addlegendentry{\(M_\otimes\)}
        
    \end{axis}
    \end{tikzpicture}
    }
    \\(a) Flickr30k
  \end{subfigure}
  \begin{subfigure}[t]{0.8\textwidth}
    \centering
    \resizebox{0.99\textwidth}{!}{ 
    \begin{tikzpicture}
    \begin{axis}[
        title={},
        width=6cm,height=5cm,
        xlabel={Key pruning rate (\%)},
        ylabel={CIDEr-D score (\%)},
        xmin=0, xmax=100,
        ymin=0, ymax=100,
        legend pos=north east,
        ymajorgrids=true,
        grid style=dashed,
    ]
    
    \addplot[color=red,mark=,]
        coordinates {
        (0,91.4)(20,80.2)(40,43.5)(60,10.5)(80,5.9)(90,5)(100,4.4)
        };
        \addlegendentry{\(M_\oplus\)}
    \addplot[color=cyan,mark=,]
        coordinates {
        (0,91.6)(20,90.3)(40,88.9)(60,82.8)(80,71.9)(90,66.6)(100,50.2)
        };
        \addlegendentry{\(M_\otimes\)}
        
    \end{axis}
    \end{tikzpicture}
    \begin{tikzpicture}
    \begin{axis}[
        title={},
        width=6cm,height=5cm,
        xlabel={Key pruning rate (\%)},
        ylabel={Signature detection rate (\%)},
        xmin=0, xmax=100,
        ymin=0, ymax=110,
        legend pos=north east,
        ymajorgrids=true,
        grid style=dashed,
    ]
    
    \addplot[color=red,mark=,]
        coordinates {
        (0,100)(20,100)(40,100)(60,99.99)(80,99.9)(90,99.6)(100,98.8)
        };
        \addlegendentry{\(M_\oplus\)}
    \addplot[color=cyan,mark=,]
        coordinates {
        (0,99.99)(20,100)(40,100)(60,100)(80,99.73)(90,99.64)(100,99.48)
        };
        \addlegendentry{\(M_\otimes\)}
        
    \end{axis}
    \end{tikzpicture}
    }
    \\(a) MS-COCO
  \end{subfigure}
  
  \caption{Removal attack (Key Pruning): CIDEr-D score and signature detection rate of our approaches on both MS-COCO and Flickr30k against different key pruning rates.}
  \label{fig:removal-key-pruning}
\end{figure*}
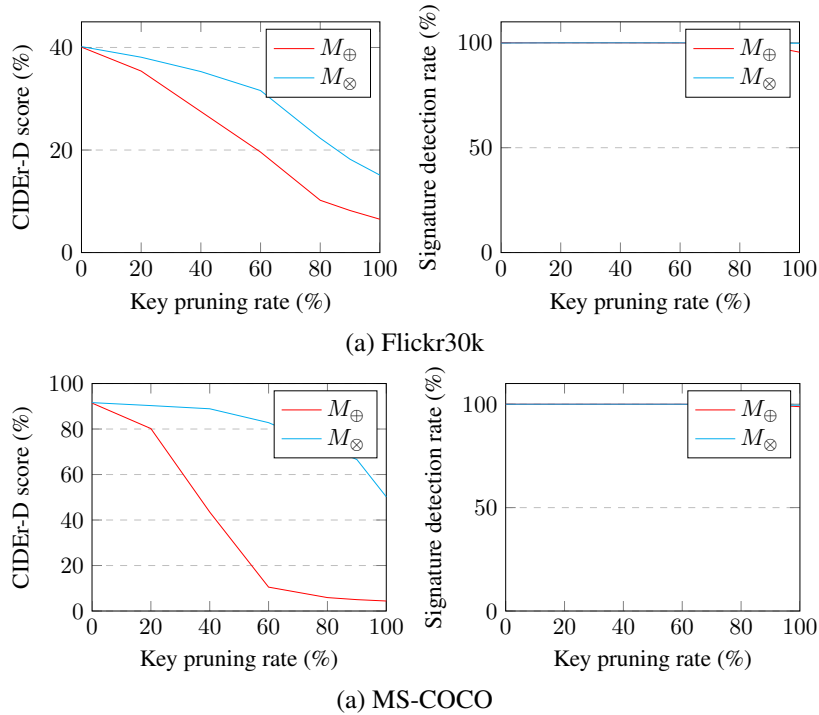

If an attacker knows a key is in place, instead of pruning the model, we simulate an attacker to prune the key rather than model weights. Figure \ref{fig:removal-key-pruning} shows the CIDEr-D score and signature detection rate on \(M_\oplus\) and \(M_\otimes\) against different key pruning rates. We show that even the attacker prunes 100\% on the key, the signature detection rate still remains at more than 98\% and 95\% on both MS-COCO and Flickr30k datasets. Since key pruning has changed the original key, it degrades the performance of the proposed model as well. As shown in Figure \ref{fig:removal-key-pruning}, the CIDEr-d score continues to drop when the key pruning rate increases. Hence, the proposed secret key and signature can protect the ownership of the model against the key pruning attack.

\subsubsection{Fine-tuning key and signature}
\label{subsubsec: fine-tuning-key-sign}
This experiment is different from Section \ref{subsubsec: fine-tuning}, here, we simulate an attacker who knows everything about the model, i.e., the training procedure, the training parameters, the dataset used, and the key/signature. The attacker fine-tunes the model with a different key/signature following the same training steps. Table \ref{tab:removal-finetune-keysign} shows the CIDEr-D score and the signature detection rate of the proposed model after perform key and signature fine-tuning. After fine-tuning the model to a new key and signature, we show that our approaches achieve a slightly lower CIDEr-D score (-1\% to -5\%) as compared to the protected model. However, the signature detection rate dropped from almost 100\% to around 68\%. This is the worst-case scenario where the model is difficult to protect against the attacker who knows everything about the model including the training steps. 

\begin{table}[t]
\centering
\resizebox{0.6\textwidth}{!}{ 
\renewcommand*{\arraystretch}{1.1}
\Huge
\begin{tabular}{|l|c|c|c|c|}
\hline
\multirow{2}{*}{Methods} & \multicolumn{2}{c|}{MS-COCO} & \multicolumn{2}{c|}{Flickr30k}\\ \cline{2-5} 
  & Protect & Attack & Protect & Attack \\ \hline
\(M_\oplus\) & 100 (91.40) &  68.08 (89.60) & 100 (40.07) & 69.14 (39.70)\\
\(M_\otimes\) & 99.99 (91.60) & 68.16 (89.6) & 99.99 (40.17) & 67.96 (38.1)\\
\hline
\end{tabular}
}
   \caption{Fine-tuning key and signature attack: CIDEr-D (in-bracket) of proposed models (Left: MS-COCO. Right: Flickr30k.) Accuracy (\%) outside bracket is the signature detection rate.}
  \label{tab:removal-finetune-keysign}
\end{table}

\subsection{Network Complexity}
\label{complexity}

\begin{table}[h]
\centering
\resizebox{0.7\textwidth}{!}{ 
\renewcommand*{\arraystretch}{1.1}
\Huge
\begin{tabular}{|l|c|c|c|c|}
\hline
\multirow{2}{*}{Methods} & \multicolumn{2}{c|}{MSCOCO} & \multicolumn{2}{c|}{Flickr30k}\\ \cline{2-5} 
  & Training time & Inference time & Training time & Inference time \\ \hline
Baseline & 20h 10m & 10.60s & 4h 25m & 10.64s \\ 
\(M_\oplus\) & 20h 31m & 10.58s & 4h 30m & 10.60s \\ 
\(M_\otimes\) & 20h 31m  & 10.45s & 4h 30m & 10.44s \\
\hline
\end{tabular}
}
   \caption{Comparison of network complexity of our approaches and baseline model. Training time is calculated over the entire training set for 20 epochs. Inference time is calculated over a single iteration, where h, m, s are hour, minute, and second.}
  \label{tab:netwcomplex}
\end{table}

Our approach with the key and signature embedding in the image captioning model does not cause the extra cost. We conducted an experiment to compare the training and inference time between baseline, $M_\oplus$ and $M_\otimes$. All the experiments used TITAN V GPU with the same setting and hyperparameter as stated in Section \ref{experiment}. On both Flickr30k and MS-COCO datasets, the complexity of our approach (with the key and signature embedded) compared to the baseline model is almost negligible. Herein, we show the complete results in Table \ref{tab:netwcomplex} to compare the training time and inference time. It is noticed that our approaches only have an incremental of 1.89\% or below in the training time on Flickr30k and MS-COCO datasets. For the inference time, it is almost similar to the baseline model for a single iteration. As a summary, we show that our proposed model provides reliable, preventive, and timely IP protection at virtually no extra cost for image captioning tasks.

\subsection{Comparison with different types of image captioning frameworks}
\label{compareother}

Despite showing our approach in the popular image captioning framework \emph{Show, Attend and Tell} model \cite{xu2015show}. We further applied our approach in the different types of image captioning frameworks, which are Up-Down \cite{anderson2018bottom} and SCST \cite{rennie2017self}. Up-Down model used the bottom-up attention techniques to find the most relevant regions based on bounding boxes and two LSTM layers are used to selectively attend to the image features to generate the caption. While SCST model applied the reinforcement learning method to image captioning by optimizing the model directly on those objective evaluation metrics like CIDEr score. Both Up-Down and SCST are having different frameworks compared to our baseline. Therefore, we need to adapt our proposed key embedding approach to the frameworks.

Up-Down model consists of two LSTM layers, the first LSTM layer as a top-down visual attention model and the second LSTM layer as a language model. It has an attention module in between the first and second LSTM layers. We apply our proposed key embedding approach in the second LSTM layer and the attention module. The key embedding process is defined as:
\begin{equation}
    h^2_{t-1} = C \cdot (h^2_{t-1} \oplus K) + (1 - C) \cdot (h^2_{t-1})
\end{equation}
\begin{equation}
    \hat{v}_t = C \cdot (\hat{v}_t) + (1 - C) \cdot (\hat{v}_t \oplus K)
\end{equation}
where $h^2_{t-1}$ is the previous hidden state of the second LSTM layer, $\hat{v}_t$ is the attended image features and $C \in \{0, 1\}$, $0$ indicates forged key, $1$ indicates real key. 

SCST model modified the architecture of the attention model for captioning given in \cite{xu2015show}, and input the attention-derived image feature only to the cell node of the LSTM. We apply our proposed key embedding approach in the attention-derived image feature and the hidden state. The key embedding process is defined as:
\begin{equation}
    h_{t-1} = C \cdot (h_{t-1} \oplus K) + (1 - C) \cdot (h_{t-1})
\end{equation}
\begin{equation}
    I_t = C \cdot (I_t) + (1 - C) \cdot (I_t \oplus K)
\end{equation}
where $h_{t-1}$ is the previous hidden state of the LSTM layer, $I_t$ is the attention-derived image feature. 

Table \ref{tab:updown} shows the performance of our approach with the Up-Down model on the MS-COCO dataset in all 5 image captioning metrics. UD-$M_{\oplus}$ refers to our proposed key embedding approach in the Up-Down model. While UD-$\widehat{M_{\oplus}}$ is similar to UD-$M_{\oplus}$ but with the forged secret key. We can observe that UD-$M_{\oplus}$ is having a lower score compared to the original Up-Down model. With the forged secret key, the performance of the model degrades significantly which showing our approach can protect the Up-Down captioning model. For example, the CIDEr-D score drops 17.27\% from 101.93 to 84.33.

According to Table \ref{tab:scst}, it shows the performance of our approach with the SCST model on the MS-COCO dataset. We follow the experiment in the original paper \cite{rennie2017self} to evaluate the model in 4 image captioning metrics: BLEU-4, METEOR, ROUGE-L, and CIDEr-D. SCST-$M_{\oplus}$ refers to our proposed key embedding approach in the SCST model. While SCST-$\widehat{M_{\oplus}}$ is similar to SCST-$M_{\oplus}$ but with the forged secret key. Our proposed key embedding approach SCST-$M_{\oplus}$ does perform poorly compared to the original SCST model but is able to protect the model against the forged secret key. With the forged secret key, the CIDEr-D score drops from 101.87 to 90.53.

\begin{table}[t]
\centering
\resizebox{0.7\textwidth}{!}{ 
\renewcommand*{\arraystretch}{1.1}
\begin{tabular}{|c|cccccc|}
\hline
\multirow{2}{*}{Methods} & \multicolumn{6}{c|}{Evaluation Metric} \\ \cline{2-7} 
  & B-1 & B-4 & M & R & C & S \\ \hline
Up-Down \cite{anderson2018bottom} & 76.97 & 36.03 & 26.67 & 56.03 & 111.13 & 19.90 \\ 
\hline
UD-$M_{\oplus}$ & 71.57 & 33.83 & 25.33 & 52.43 & 101.93 & 18.60 \\
UD-$\widehat{M_{\oplus}}$ & 65.20 & 29.50 & 20.33 & 48.60 & 84.33 & 16.53 \\
\hline
\end{tabular}
}
  \caption{Comparison between our approach (UD-$M_{\oplus}$) with Up-Down \cite{anderson2018bottom} model on MS-COCO dataset, across 5 common metrics where B-N, M, R, C and S are BLEU-N, METEOR, ROUGE-L, CIDEr-D and SPICE scores. UD-$\widehat{M_{\oplus}}$ is similar to UD-$M_{\oplus}$ but with the forged secret key.}
  \label{tab:updown}
\end{table}

\begin{table}[t]
\centering
\resizebox{0.6\textwidth}{!}{ 
\renewcommand*{\arraystretch}{1.1}
\scriptsize
\begin{tabular}{|c|cccc|}
\hline
\multirow{2}{*}{Methods} & \multicolumn{4}{c|}{Evaluation Metric} \\ \cline{2-5} 
  & B-4 & M & R & C \\ \hline
SCST \cite{rennie2017self} & 33.87 & 26.27 & 55.23 & 111.33 \\
\hline
SCST-$M_{\oplus}$ & 31.60 & 24.97 & 52.43 & 101.87 \\
SCST-$\widehat{M_{\oplus}}$ & 29.83	& 22.63 & 50.17 & 90.53 \\
\hline
\end{tabular}
}
  \caption{Comparison between our approach (SCST-$M_{\oplus}$) with SCST \cite{rennie2017self} model on MS-COCO dataset. SCST-$\widehat{M_{\oplus}}$ is similar to SCST-$M_{\oplus}$ but with the forged secret key.}
  \label{tab:scst}
\end{table}

\section{Limitations}
\label{limitations}
We show that our proposed secret key together with signature can protect the model against unauthorized usage as simulated in Section \ref{experiment}. However, the embedding-based methods have some limitations that are inevitable. For example, the secret key can be removed and the signature detection rate is dropped in the worst-case scenario in which the attacker knows everything about the model as shown in Section \ref{subsubsec: fine-tuning-key-sign}. This hinders open-sourcing the model as we need to prevent others from knowing the secret key, the training procedure, and the exact training parameters.

\section{Conclusion}
\label{conclusion}

IP protection on DNN has been a significant research area and we take the first step to implement the ownership protection on the image captioning task. The protection is achieved in two different embedding schemes, using the hidden memory state of RNN so that the image captioning functionalities are paralyzed for unauthorized usage. We demonstrated with extensive experiments that our proposed, on the one hand, the image captioning functionalities are well-preserved in the presence of valid secret key and well-protected for unauthorized usages on the other hand. The proposed key-based protection is, therefore, more cost-effective, proactive, and timely, as compared with watermarking-based protections which have to rely on government investigation and juridical enforcing actions. However, the proposed key-based protection also has some weaknesses, where the overall protection will be compromised when the attacker knows everything about the model. This is the direction in our future work to solve these weaknesses and ensure the model can be fully protected against different types of attackers.

\subsection{Broader Impact}
\label{impact}

Our work is mainly focused on protecting the image captioning model with ownership verification. The engineer or researcher of the image captioning model might benefit from this research to protect their model against IP infringement. This is crucial as the development and training of an image captioning model is expensive especially when it involves a very large dataset. IP protection in image captioning is critical to fostering innovation in this field to achieve top-level performance that can benefit the society. We believe that no one in genuine may be put at disadvantage from this work. In case of the failure of our work in protecting the model, the worst scenario is an attacker can access the model without owner acknowledgement. In short, our work is bringing benefit to the society especially to AI start-ups to secure their advantage in the open market. We will also make the source code of this work publicly available for people to reproduce and follow up.

\bibliographystyle{IEEEbib}
\bibliography{refs}

\end{document}